\pdfoutput=1
\documentclass[sigconf]{acmart}

\usepackage{pifont}
\usepackage{multirow}
\usepackage{adjustbox}
\usepackage{color}
\usepackage{threeparttable}
\usepackage{makecell}\usepackage{multirow}
\usepackage{color}
\usepackage{mathtools}
\usepackage{subfigure}
\usepackage{makecell}
\usepackage{adjustbox}
\usepackage{float}

\usepackage{tikz}
\usetikzlibrary{patterns}

\usepackage{pgfplots}
\pgfplotsset{compat=1.18}

\AtBeginDocument{%
  }

\setcopyright{acmlicensed}
\copyrightyear{2025}
\acmYear{2025}
\acmDOI{XXXXXXX.XXXXXXX}
\acmConference[KDD '25]{August 03--07,
  2025}{Toronto, Canada}

\begin{document}
\title{Strada-LLM: Graph LLM for traffic prediction}

\author{Seyed Mohamad Moghadas}
\affiliation{%
  \institution{Department of Electronics and Informatics, Vrije Universiteit Brussel}
  \city{B-1050 Brussels}
  \country{Belgium}
  }
\email{Seyed.Mohamad.Moghadas@vub.be}
\orcid{0009-0006-0950-9315}

\author{Bruno Cornelis}
\additionalaffiliation{%
  \institution{Macq}
  \city{B-1140 Brussels}
  \country{Belgium}
}
\affiliation{%
  \institution{Department of Electronics and Informatics, Vrije Universiteit Brussel}
  \city{B-1050 Brussels}
  \country{Belgium}
}
% \affiliation{%
%   \institution{MACQ}}

\email{bcorneli@etrovub.be}
\orcid{0000-0002-0688-8173}

\author{Alexandre Alahi}
\additionalaffiliation{%
  \institution{VITA Lab, EPFL}
  \city{Lausanne}
  \country{Switzerland}
}
\affiliation{%
  \institution{VITA Lab, EPFL}
  \city{Lausanne}
  \country{Switzerland}
}
% \affiliation{%
%   \institution{MACQ}}

\email{alexandre.alahi@epfl.ch}
\orcid{0000-0002-5004-1498}

\author{Adrian Munteanu}
\authornote{Member, IEEE}
\affiliation{%
  \institution{Department of Electronics and Informatics, Vrije Universiteit Brussel}
  \city{B-1050 Brussels}
  \country{Belgium}
}
\email{Adrian.Munteanu@vub.be}
\orcid{0000-0001-7290-0428}

% \author{Anonymous authors}
% \renewcommand{\shortauthors}{Anonymous}

\begin{abstract}
Traffic forecasting is pivotal for intelligent transportation systems, where accurate and interpretable predictions can significantly enhance operational efficiency and safety. A key challenge stems from the heterogeneity of traffic conditions across diverse locations, leading to highly varied traffic data distributions. Large language models (LLMs) show exceptional promise for few-shot learning in such dynamic and data-sparse scenarios. However, existing LLM-based solutions often rely on prompt-tuning, which can struggle to fully capture complex graph relationships and spatiotemporal dependencies—thereby limiting adaptability and interpretability in real-world traffic networks.

We address these gaps by introducing Strada-LLM, a novel multivariate probabilistic forecasting LLM that explicitly models both temporal and spatial traffic patterns. By incorporating proximal traffic information as covariates, Strada-LLM more effectively captures local variations and outperforms prompt-based existing LLMs. To further enhance adaptability, we propose a lightweight distribution-derived strategy for domain adaptation, enabling parameter-efficient model updates when encountering new data distributions or altered network topologies—even under few-shot constraints.

Empirical evaluations on spatio-temporal transportation datasets demonstrate that Strada-LLM consistently surpasses state-of-the-art LLM-driven and traditional GNN-based predictors. Specifically, it improves long-term forecasting by 17\% in RMSE error and 16\% more efficiency. Moreover, it maintains robust performance across different LLM backbones with minimal degradation, making it a versatile and powerful solution for real-world traffic prediction tasks.
\end{abstract}


\maketitle

\section{Introduction}

With the development of Intelligent Transportation Systems, spatio-temporal traffic prediction has received increasing attention. It is a key component of advanced urban management systems and is crucial in urban planning, mobility management, and resource allocation. Spatio-temporal prediction involves analyzing urban conditions on various dimensions, including flow, speed, and density, mining their patterns, and predicting trends ~\cite{li2018diffusion}. This capability provides a scientific foundation for any urban management department to anticipate and mitigate congestion, implement preemptive restrictions, and enable urban residents to select safer and more efficient travel routes.

\begin{figure}[!htb]
\includegraphics[width=\linewidth,keepaspectratio]{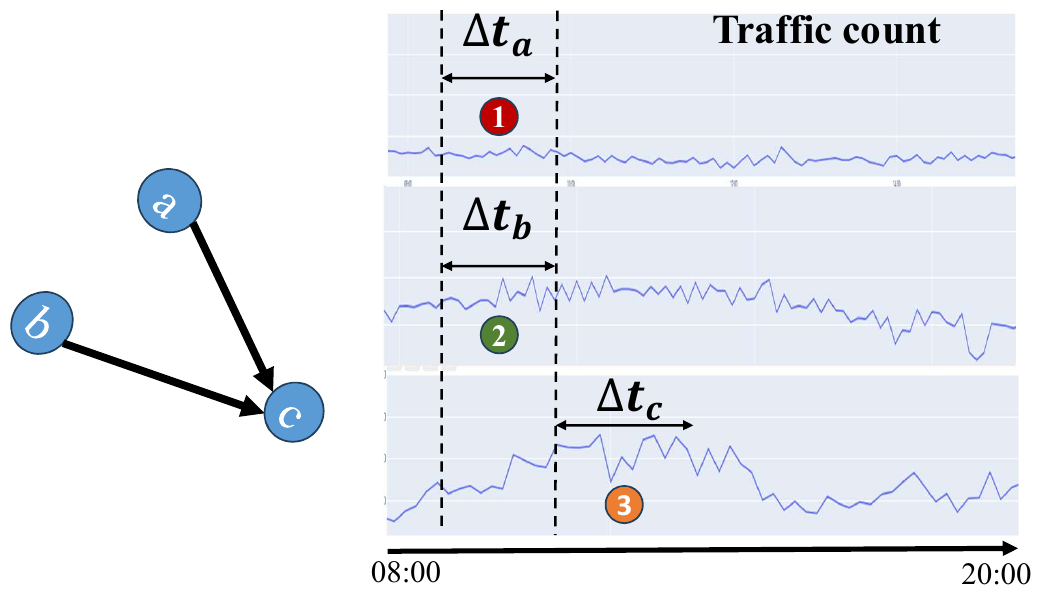}
% \vspace{-1cm}
\centering\caption{An example of the spatio-temporal dependency between the traffic flows for different traffic states \ding{172},\ding{173},\ding{174}. Nodes (a), (b), and (c) are placed in a crowded area of Brussels. The corresponding traffic patterns between 8:00 am and 8:00 pm are illustrated for nodes (a), (b), and (c). The short-term traffic flow in (c) is altered by the significant impact of neighboring roads, resulting in varying flows.}
% \centering\caption{proposed LoRA method for transfer learning}
\label{figintro}
\end{figure}

However, spatio-temporal prediction is a challenging task due to complex spatial and temporal dependencies:

1) Spatial Dependency: The variation in urban patterns is influenced by the topological structure of the urban network. In particular, conditions at upstream locations impact downstream locations through transfer effects, which refers to the impact that a change in one section (e.g., a closure of an area, a change in service timing, or new infrastructure) has on other parts of the network. On the other hand, downstream locations affect upstream locations through feedback effects, which occur when the changes in patterns resulting from transfer effects influence the original cause of the change, creating a loop of cause and effect.

2) Temporal Dependency: Urban conditions change over time, exhibiting periodicity and trends, which are often influenced by factors such as holidays, working hours, and other social events. As illustrated in Figure ~\ref{figintro}, the strong influence among adjacent locations caused by the flow alongside the edges changes the short-term states. Indeed, location (a) has a steady count trend, and location (b) has short spikes in volume, so the spatio-temporal correlation yielded a different state in location (c) after a couple of hours. Furthermore, urban data sources exhibit a variety of data distributions, posing a significant challenge for finding generalizable models.

In recent years, the emergence of Large Language Models (LLM) has revolutionized various fields due to the generalization capabilities of extracting efficient features from different modalities ~\cite{unist}. Research has been conducted in multiple domains, such as computer vision and natural language processing. Specifically, in ~\cite{unist}, a unified representation space is proposed, namely shared vocabulary, across varied data sources such that the model can learn domain-invariant features. This characteristic is particularly appealing for urban prediction, where data distributions can vary significantly across different regions and times.

Previous works ~\cite{urbangpt, li2024flashstsimpleuniversalprompttuning} have tried to address this problem by casting the graph (and/or the corresponding time-series) to a prompt input. However, this branch of methods, in terms of predictability, is suboptimal because traditional time series forecasting models use numerical inputs and outputs, whereas prompt-based methods transform these into text prompts, which may not be as effective for capturing the characteristics and trends of time series ~\cite{rasul2024lagllamafoundationmodelsprobabilistic}. It leads to a performance gap compared to classical Graph Neural Networks ~\cite{10.1145/3534678.3539396}. Moreover, the complexity and scalability of the prompt-based approaches are unexplored. For instance, in Spatio-Temporal Graphs (STG) with 1 million nodes, feeding the nodes to an LLM with limited context length is a challenging operation. Consequently, the prompting operation loses the global topological information.

% Spatial reasoning plays a pivotal role when combining the physical layout of data with temporal analysis, as it allows models to pinpoint and interpret complex relationships that extend beyond mere time-series snapshots. In contexts such as urban monitoring, understanding spatial relationships—like connectivity between nodes or correlations between geographic regions—enables a richer perspective on evolving phenomena. By accounting for both local adjacency (e.g., immediate neighbors in a network) and global structure (e.g., entire regional patterns), models can better capture how events in one location reverberate throughout the system.

Effective LLM adaptation approach have a noticeable impact on their Zero-Shot performance. There are three prominent approaches to LLM adaptation\cite{alghisi-etal-2024-fine-tune}. Namely, Prompt-Tuning\cite{qian-etal-2023-harnessing}, Adapter-Layer\cite{alghisi-etal-2024-fine-tune}, Parameter-Efficient Fine-Tuning (PEFT)\cite{hu2021lora}. The adapter-layer approach~\cite{alghisi-etal-2024-fine-tune} tends to be more applicable in the multi-task setup for foundational models\cite{alghisi-etal-2024-fine-tune}. In contrast, the effectiveness of Prompt-based approaches and PEFT remains a point of contention in the literature. Both methods have emerged as favored for integrating domain-specific expertise into Large Language Models (LLMs). However, in the realms of existing urban LLMs, UrbanGPT\cite{urbangpt} and UniST\cite{unist} have focused on Prompt-based approaches. Specifically, they include a \textbf{learnable} prompt network (PN) into the LLM. Essentially, the PN is fine-tuned in the adaptation stage. Prompt networks could impose extra complexity and neglect scalable features in the spatio-temporal target domain. Specifically, PN relies on the memory network\cite{NIPS2015_8fb21ee7}, which has been proven in graph-based spatio-temporal data to cause exponential error propagation\cite{barbero2024localityawaregraphrewiringgnns}. On the other hand, PEFT\cite{hu2021lora} involves modifying a limited set of model parameters—either by targeting existing weights or introducing additional ones—while preserving the bulk of the model's architecture. PEFT enables the retention of valuable pre-trained feature representations while concurrently tailoring the model to particular domains through targeted adjustments. However, the competence and scalability of PEFT in the LLM-based urban traffic forecasting domain remain unexplored. 

To address these issues, we propose a new forecasting method called Strada-LLM, which is a graph-aware LLM for urban traffic prediction. To the best of our knowledge, our model is the first to propose a specialized parameter-efficient LLM model in traffic forecasting. Our contributions are five-fold:

\begin{enumerate}
    \item Strada-LLM is a probabilistic LLM compliant with classical LLMs and specializes in spatio-temporal prediction. Specifically, Strada-LLM has the capability to predict urban traffic patterns, even when faced with a data distribution that differs from its training data, particularly in the few-shot setting.
    \item To the best of our knowledge, Strada-LLM is the first non-prompt-based probabilistic LLM that has lightweight adaptation capability, effectively capturing both local and global spatial dependencies.
    \item Strada-LLM is graph-aware, taking spatial dependencies into account by encoding the network graph implicitly and without any prompting. This allows the model to maintain crucial topological information throughout the prediction process.
    \item Strada-LLM adopts a lightweight approach to perform domain adaptation. Specifically, adopting low-ranked approaches showcases its effectiveness in adapting to new datasets while maintaining computational efficiency.
    \item We provide a comprehensive evaluation and demonstrate superior accuracy compared to existing urban LLMs. We evaluate our approach using numerous real-world datasets. Our results show a reduction in prediction error, ranging from approximately 5\% to 18\%, compared to baseline methods. This demonstrates the superiority of our graph-aware LLM model in urban forecasting.
    % \item Strada-LLM is flexible and can be used with different known LLMs, making it adaptable to various model architectures and requirements.
\end{enumerate}
The remainder of this paper is organized as follows: Section 2 presents the preliminaries, Section 3 describes the methodology, Section 4 discusses the results, and Section 5 concludes.
\section{Preliminaries}
\textbf{Spatio-Temporal Forecasting}. 
In this paper, we aim to predict traffic conditions at a specific timestamp $T{'}$ based on historical traffic data over period $T$. Particularly, the traffic metric broadly refers to traffic speed, flow, or density.
% Our approach defines traffic information broadly, encompassing traffic speed, flow, or density. For illustrative purposes in the experiment section, we use traffic speed as a representative example.
Before further elaborating on our method, we define the following notations. In Strada-LLM, a road network is represented as a graph $G$, consisting of a set of nodes $V$ and a set of edges $E$. Each node presents a traffic measuring sensor, and an edge connects two nodes if they are geographically adjacent. That is, $G = (V, E, A)$ where $V$ is a set of $N$ nodes, $E$ is a set of edges and $A \in \mathbb{R}^{N \times N}$ is the corresponding adjacency matrix. We assume that the topology of the traffic graph $G$ is static. At each timestamp $t$, the graph $G$ is associated with a dynamic feature matrix $\boldsymbol{X}_t \in \mathbb{R}^{N \times F}$, where $F$ is the number of traffic metrics.
% \textcolor{red}{In this paper, we consider traffic speed.}
% , which is traffic speed in this paper.

% A traffic road network is represented as a graph $G$ at timestamp $t$, consisting of a set of nodes $V$ and a set of edges $E$. Each node presents a traffic measuring sensor and an edge connects two nodes if they are geographically adjacent; their distance is lower than a certain threshold, chosen empirically. So, the graph is defined as $G = (V, E, A)$ where $V = \left\{V_1, V_2, ..., V_N\right\}$ is a set of nodes, $E$ is a set of edges and $A \in \mathbb{R}^{N \times N}$ is the adjacency matrix. At each timestamp, $t$, the graph $G$ contains a dynamic feature matrix $\boldsymbol{X}_t \in \mathbb{R}^{N \times D}$, where $N$ is the number of nodes, and $D$ is the number of traffic features, which, in this case, is traffic speed. In this research, we assume that the topology of the traffic graph is static.

In this paper, we formulate traffic forecasting as a discrete multi-variate point prediction problem. In particular, Strada-LLM takes the $T$ past observations  $\boldsymbol{X} = \left[ \boldsymbol{X}_{t_1}, \boldsymbol{X}_{t_2}, . . . , \boldsymbol{X}_{t_T} \right]\in\mathbb{R}^{T \times N \times F}$ from $G$ as input and predicts the traffic for horizon $T'$, where:

${\boldsymbol{\hat{X}} = \left[\boldsymbol{\hat{X}}_{t_{T+1}}, \boldsymbol{\hat{X}}_{t_{T+2}}, ... , \boldsymbol{\hat{X}}_{t_{T+T'}} \right]}$. 

\section{Methodology}
\label{sec:solution}

Strada-LLM consists of a Hierarchical Feature Extractor (HFE), an LLM-based backbone, and a T-student distribution head, as shown in Figure~\ref{fig1}. The distribution head includes three learnable parameters related to the output: degrees of freedom, mean, and scale, to ensure the relevant parameters remain positive. In the HFE module, a sub-graph extractor and a global graph feature extractor hierarchically learn features while the LLM-based block maps the graph features into the latent spatio-temporal feature space to be learned by the distribution head. In the following sections, we will elaborate on each component.
% Figure~\ref{fig1} shows the building blocks of the proposed model with $M$ decoder layers. The model contains a sub-graph extractor module, Mistral~\cite{jiang2023mistral7b} transformer decoder, and the t-student distribution $t(\nu)$. In the following sections, each block functionality is detailed.

\subsection{Hierarchical Feature Extractor}

\subsubsection{{$k$-hop Sub-Graph Extractor:}}

To enhance the LLM's understanding of graph structures while accommodating its limited context length, we tokenize the spatio-temporal traffic signal by taking into account the existing subgraphs of the road network. Overall, the corresponding traffic signals for neighboring nodes will be aggregated together to be tokenized in the further steps. This block is responsible for extracting $k$-hop subgraphs from the input graph. For a node $v \in G$, the k-hop operator~\cite{knn} is defined as:

\begin{equation}
\label{khop}
\mathcal{N}_k(v) = \left\{ u\in V | d(u,v) \leq k \right\},
\end{equation}
where $d(\cdot, \cdot)$ is the hop-based distance function between two nodes. 
By doing so, each node is equipped with a sub-graph of $G$, including local topological information. As a result, for each node $v$, we concatenate the features of $\mathcal{N}_k(v)$, leading to an $N \times T \times (M\times F)$ feature map, where $M$ denotes $|\mathcal{N}_k(v)|$, cardinality of neighbors. Similarly, Subgraph-1-WL~\cite{pmlr-v238-zhou24a} and MixHop~\cite{abuelhaija2019mixhop} can also be used as the k-hop operator, but it is beyond the scope of this research.

\subsubsection{Global Graph Embedding}
While prompt-based networks are widely used in the literature~\cite{10.1145/3637528.3671451}, {they lose the global structure information~\cite{li2024flashstsimpleuniversalprompttuning}.} As demonstrated by ~\cite{JMLR:v22:19-683}, in order to keep the global graph structure information, we compute Laplacian embeddings. The Laplacian operation leverages the eigenvectors of the complete graph and creates unique{ positional encodings~\cite{vaswani2023attention} }for each node.
First, we define the graph Laplacian matrix\:
$\textbf{L}=D - A$
where
$D=\text{diag}(d_1,\ldots,d_N)$ is the degree matrix.
We then compute the eigendecomposition of the normalized Laplacian as follows
$L_{\text{norm}} = D^{-1/2}\textbf{L}D^{-1/2} = \textbf{U}\Lambda \textbf{U}^T$. {Then we compute global embedding by non-linear mapping of \textbf{U} matrix. Reader can refer to Section~\ref{scalability_section} to discern the contribution of this representation extraction on the LLM forecasting ability. We denote $F'$ as the embedding dimension.}   
% \textcolor{red}{where $\Lambda$ is the diagonal matrix?}

\subsubsection{Lag Extractor}
% The tokenization strategy employed by Strada-LLM revolves around the creation of lagged features derived from historical traffic data. These features are constructed using predetermined sets of lag indices, such as quarterly, monthly, etc., also referred to as second-level frequencies, learned to capture various temporal patterns.
Mathematically, this lagging operation can be expressed as a mapping $\boldsymbol{x}_{t} \longmapsto \boldsymbol{\lambda_t} \in \mathbb{R}^{|\mathcal{H}| \times F}$, where $\mathcal{H} = \left\{h_1, h_2, ..., h_H\right\}$ represents the set of historical lag indices. In this formulation, {$\lambda_t[j] = x_{t-\mathcal{H}[j]}$}, meaning that $\lambda_t[j]$ corresponds to the value of $x$ at $h_j$ time steps before $t$, as specified by the $j$-th element of $\mathcal{H}$.

In our approach, we extend beyond simple lag feature extraction by employing a structured tokenization scheme to convert spatio-temporal slices of the road network graph into discrete tokens for the LLM-based backbone. Specifically, for each time step \( t \), we first derive both global and local (k-hop) graph embeddings. These representations are then segmented within a sliding window, yielding embeddings that encapsulate the node-level or sub-graph-level features alongside their temporal contexts. Crucially, we add positional embeddings to each token so that the masked Transformer decoder can effectively attend to different time steps in a sequence-like manner. While lag features help capture short-term correlations, the tokenization scheme formalizes how each time window becomes a coherent input unit, thus ensuring the LLM’s attention mechanisms can leverage meaningful traffic patterns and dependencies across time and space.

Therefore, to generate lag features for a specific context-length window $x_{1:C}$, where $C$ is the length of the context window, it is necessary to consider an expanded window that includes $H$ additional historical data points. Besides the lagged features, we incorporate date-time features covering all temporal features in our dataset, including second-of-minute, hour-of-day, and so forth, up to quarter-of-year, based on the time index $t$. It is important to note that the main purpose of these date-time features is to enrich the information set. In any time series, all but one of the date-time features will stay the same from one-time step to another, allowing the model to intuitively understand the time series frequency. In the end, the HFE block's output is a tensor in shape $\mathbb{R}^{C \times (1+M) \times |\mathcal{H}| \times (F+F')}$.

% , leading to an informative feature matrix \textcolor{red}{TODO put the extended feature matrix notation here?} associated with local graph information. 

% Suppose that the input graph tensor dimensionality would be $\mathbb{R}^{N \times T \times D}$, after applying this operation, it transforms into $\mathbb{R}^{N \times T \times (M\times D)}$, where $M=\min_{u \in V}(|\mathcal{N}_k(u)|)$. Indeed, after extracting sub-graphs for each node, we concatenate the node traffic attributes with each other. It is worth pointing out that other approaches, such as Subgraph-1-WL \cite{pmlr-v238-zhou24a} and MixHop \cite{abuelhaija2019mixhop} could also be used, which is beyond the scope of this research.

\subsection{LLM-based-Backbone}

Our backbone architecture is inspired by {Mistral~\cite{jiang2023mistral7b}}, a decoder-only transformer architecture. The extracted tokens by the sub-graph extractor are transformed by a shared linear projection layer that maps the features to the hidden dimension of the attention module. Inspired by~\cite{touvron2023llama}, Strada-LLM includes
pre-normalization by utilizing the RMSNorm~\cite{Zhou_Zhang_Peng_Zhang_Li_Xiong_Zhang_2021} and Rotary Positional Encoding (RoPE)~\cite{su2023roformer} for the representation of query and key blocks at each attention layer, similar to the approach described for Mistral~\cite{jiang2023mistral7b}. 
% We also take Laplacian embedding into account. Laplacian positional embedding is a powerful technique to incorporate structural information about nodes' positions within a graph. This method leverages the eigenvectors of the graph Laplacian matrix to create unique positional encodings for each node, effectively capturing their relative positions and global graph structure. 

% where\:
% $U = [u_1, \ldots, u_n]$ \text{ is the matrix of eigenvectors} $U=[u1,…,un]$ is the matrix of eigenvector s$\Lambda=diag(\lambda_1,…,\lambda_n)$ is the diagonal matrix of eigenvalues. The Laplacian positional embedding for node $i$ is then given by:
% $pi=[ui1,ui2,…,uik]p_i = [u_{i1}, u_{i2}, \ldots, u_{ik}]pi=[ui1,ui2,…,uik​]$

\begin{figure*}[!t]
\centering
\includegraphics[width=0.75\textwidth, keepaspectratio]{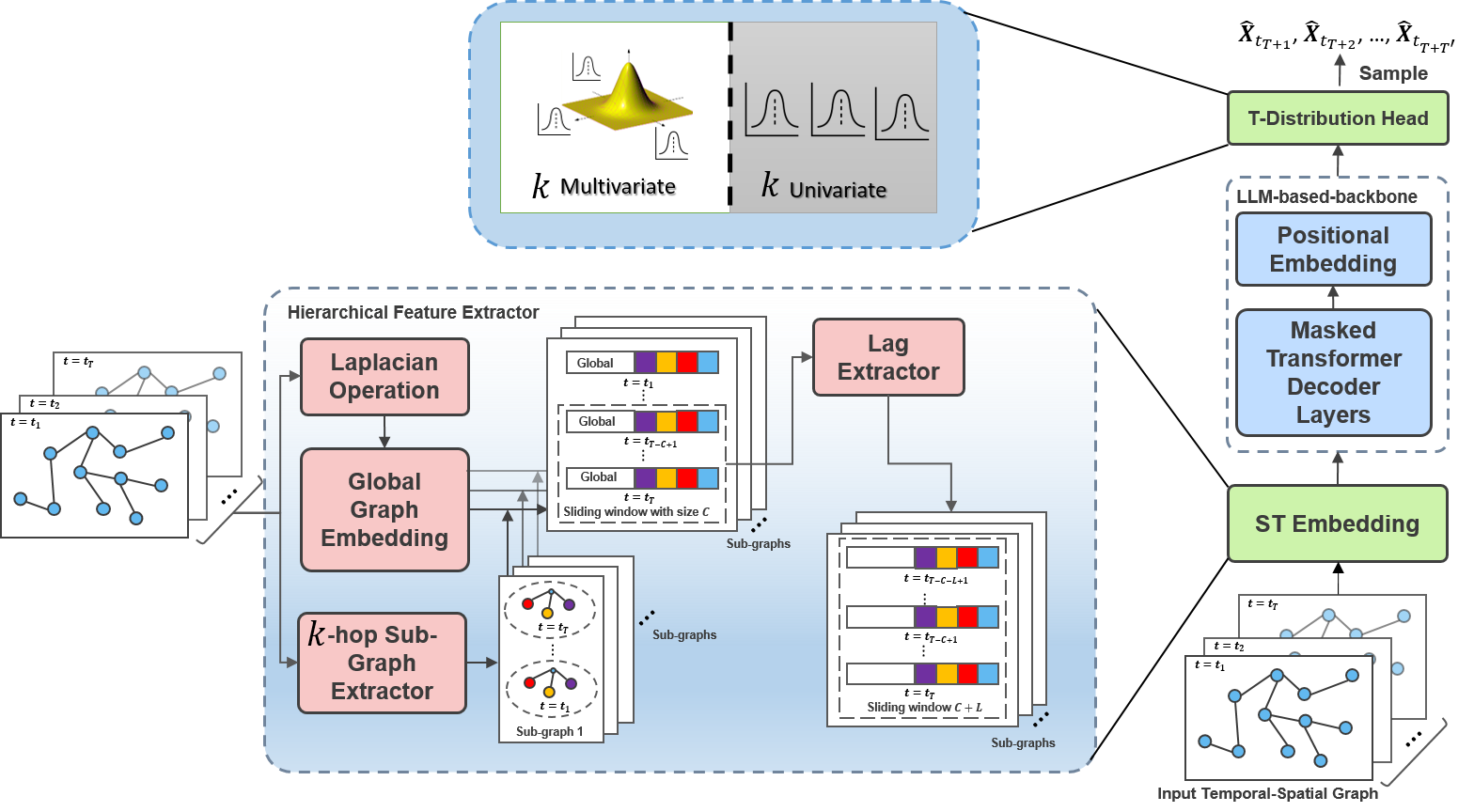}
% \vspace{-3cm}
\caption{The proposed Strada-LLM architecture. {The core novelty lies in the K-hop sub-graph extractor module, which takes lag features in both the spatial and temporal dimensions.}
% \textcolor{red}{1) remove all the colorful empty boxes, use a unique one. 2) uniqulize the style of boxes, rectangle/circle-rectangle.  3) check the inputs and outputs of each block for instance, positional embedding with addition of what? the input of Laplacian block is what etc. 4) increase the font size of formula and decrease the font size of block, make sure that both of them are readable. 5) check if Mistral is correct cited here. 6) in the masked transformer decoder layer, put something about domain adaptation.}
}
\label{fig1}
\end{figure*}

After passing through the causally masked transformer layers, the proposed method incorporates Flash-Attention-2~\cite{dao2023flashattention2fasterattentionbetter} to predict the parameters of the forecast distribution for the next timestamp, denoted as \( \phi \). These parameters are generated by a parametric distribution head, as shown in Figure~\ref{fig1}. The objective is to minimize the negative log-likelihood of the predicted distribution across all prediction times.

At the inference stage, given a time series with the size of at least {$H+C$}, a feature vector can be formed and supplied to the model to determine the distribution of the subsequent timestamps. In this fashion, we can obtain
many simulated future trajectories up to our chosen prediction horizon $T'$ via auto-regressive decoding~\cite{10.1093/oso/9780199214655.001.0001}. In doing so, uncertainty intervals and confidence scores could also be valuable for downstream decision-making tasks.

\subsection{Domain Adaptation}\label{domain}

In this section, we describe the techniques we adopt to adapt the Strada-LLM model to a new dataset, which might contain a graph of a new city. One of the important aspects of such a model is that it should be domain-agnostic. In the rapidly evolving landscape of deep learning, domain adaptation has emerged as a crucial technique for enhancing the generalizability of models across different but related domains. {More specifically, domain adaptation} focuses on the ability of a model trained in one domain (source) to adapt and perform well in another domain (target) with potentially different distributions. A prominent approach within this realm is low-rank adaptation~\cite{hu2021lora}, which aims to refine pre-trained models by adapting their parameters in a computationally efficient manner; indeed, a nearly linear operation will replace the $\mathcal{O}(n^2)$ operation. This technique is particularly beneficial when working with LLMs, which, despite their impressive performance and versatility, pose significant challenges regarding computational resources and training time. In this paper, we adapt to the new distribution by tuning the distribution head, query, key, and value attention blocks.

\begin{figure}[!t]
    \centering
    \subfigure[Strada-LLM (our method)]{
        \centering
        \includegraphics[width=0.22\textwidth,keepaspectratio]{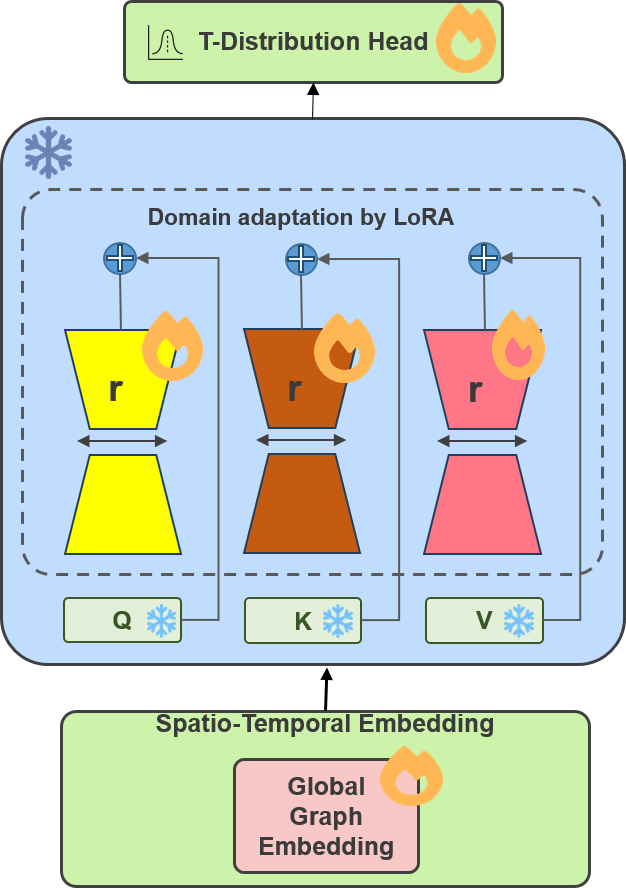}
        % \caption{Strada-LLM}
        \label{fig-2-a}
    }
    % \hfill
    \subfigure[UniST\cite{unist}]{
        \centering
        \includegraphics[width=0.2\textwidth,keepaspectratio]{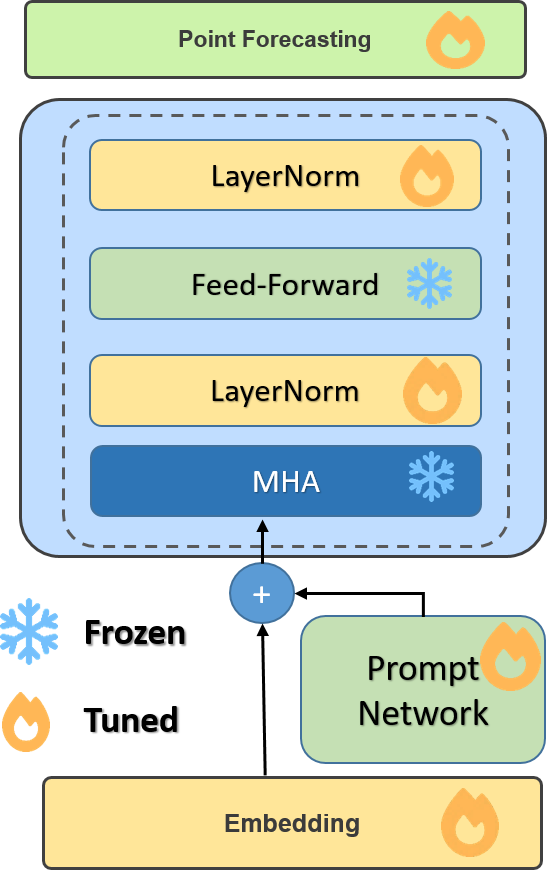}
        % \caption{UniST\cite{unist}}
        \label{fig-2-b}
    }
    \caption{Conceptual LLM adaptation approaches comparison}
    \label{fig:fig2}
\end{figure}

\subsubsection{{Low-Rank Matrix Adaptation}}
{By leveraging domain adaptation strategies and Low-Rank Matrix Adaptation (LoRA) for fine-tuning,} {researchers and practitioners can effectively bridge the gap between domains, ensuring that powerful deep learning models maintain their efficacy across diverse applications~\cite{filatov2023low}.} One of the purposes of this research is to propose a LoRA-based method for the temporal graph forecasting problem. As illustrated in Figure~\ref{fig:fig2}, we apply LoRA for the LLM fine-tuning process such as spatio-temporal patterns of the target data being learned. The query, key, and value matrices will be fine-tuned in a low-rank approximation manner. Specifically, according to Figure~\ref{fig:fig2}, suppose the query, key, and value tensors are computed as follows:
\begin{equation}
    \small
    \label{lora_pre}
    q = W_{q} h,
    k = W_{k} h,
    v = W_{v} h,
\end{equation}
where $h$ is the output of the last layer of the LLM backbone with dimension $\mathbb{R}^{d \times k}$. $W_{q}, W_{k}$ and $W_{v}$ are the query, key, and the projection weight matrices respectively. The low-rank matrices for fine-tuning the forward step can be defined as:
\begin{equation}
\small
\begin{split}
    \label{lora}
    \small
    q = W_{q} h + \Delta W_qh =  W_{q}  h + B_qA_qh \\
    k = W_{k} h + \Delta W_kh =  W_{k}  h + B_kA_kh \\
    v = W_{v} h + \Delta W_vh =  W_{v}  h + B_vA_vh
\end{split}
\end{equation}
where $h$ is the output of the previous layer, the low-ranked approximated update matrices $B_q, B_k, B_v$ are of dimensions $\mathbb{R}^{d \times r}$,  $A_q, A_k, A_v$ are of dimensions $\mathbb{R}^{r \times k}$, and where the rank hyperparameter {$r \ll min(d, k)$.} By introducing these matrices, online inference latency reduction is achieved~\cite{hu2021lora}. For the initialization of the matrices $A$ we choose Gaussian initialization while the  $B$ matrices are initialized to zero~\cite{hu2021lora}. Following this approach enables us to perform domain adaptation for different datasets consisting of diverse distributions. {As shown in Figure~\ref{fig-2-a}, our adaptation approach is conceptually different than prompt-based models like UniST~\cite{unist}. Based on the distribution-driven alignment loss function in the adaptation phase, global graph embedding, query, key, value projections, and distribution head are targeted to be tuned in the low-rank fashion. As reader can see in the experiment section~\ref{sec_comp},} tuning the mentioned blocks is more lightweight than tuning prompt-based models like Figure~\ref{fig-2-b}. {One might critique not applying low-rank tuning for other modules like Global Graph Embedding in Figure~\ref{fig-2-a}, which is addressed in Appendix ~\ref{grad_vis}.}

Recent theoretical work shows that domain alignment can be bounded using the optimal transport distance functions~\cite{yao2020learningmultiplecitiesmetalearning}. For instance, unsupervised distance adaptation approaches~\cite{ramesh2024grouprobustpreferenceoptimization} have been adopted in modern LLMs. As shown in Figure ~\ref{fig-2-a}, In Strada-LLM, the distribution head will be tuned. So, as another novel pillar of our model, during the adaptation phase, we minimize Kullback–Leibler divergence $D_{KL}(.)$ distance between multivariate distribution heads. As a matter of tractability, given a bounded $k$-variate $t$-distribution, the distance function can be derived as:
$$
\small{
\begin{aligned}
& D_{KL}(f_d(\cdot|\mu,\Sigma,\nu)||f_d(\cdot|\mu,\Sigma,\nu')) = \\
& = D_{KL}(f_d(\cdot|0,\mathbf{I},\nu)||f_d(\cdot|0,\mathbf{I},\nu')) \\
& = \int_{\mathbb{R}^n} f_d(\mathbf{x}|0,\mathbf{I},\nu)\log\frac{f_d(\mathbf{x}|0,\mathbf{I},\nu)}{f_d(\mathbf{x}|0,\mathbf{I},\nu')} d\mathbf{x}
\end{aligned}}
$$
Inspired by ~\cite{villa2018objectivepriorsnumberdegrees}, this Kullback Leibler divergence can be written as: 
\begin{equation}
\begin{aligned}
& D_{KL}(f_d(\cdot|\mu,\Sigma,\nu)||f_d(\cdot|\mu,\Sigma,\nu')) = \\
& \log\frac{K(d,\nu)}{K(d,\nu')} - \frac{\nu+d}{2}\left[\Psi\left(\frac{\nu+d}{2}\right) - \Psi\left(\frac{\nu}{2}\right)\right] + \\
& \frac{\nu'+d}{2}K(d,\nu)\frac{\pi^{\frac{d}{2}}}{\Gamma(\frac{d}{2})}\int_0^{\infty}\left(1+\frac{t}{\nu}\right)^{-\frac{\nu+d}{2}}t^{\frac{d}{2}-1}\log\left(1+\frac{t}{\nu'}\right)dt\\
& \coloneqq{\mathcal{L}_{alignment}}
& 
\label{kl_formula}
\end{aligned}    
\end{equation}
where, $\small{K(d,\nu) = \frac{\Gamma(\frac{\nu+d}{2})}{\Gamma(\frac{\nu}{2})\sqrt{(\pi\nu)^d}}}$. The proof unfolds in Appendix \ref{kl_proof}.

\subsection{Loss Function}
The Negative Log-Likelihood (NLL) is a loss function commonly used in probabilistic time-series forecasting. It is calculated as the negative of the log of the probability of the true value given the predicted value. The formula for negative log-likelihood is:

\begin{equation}
    \small
    \label{nll}
    {\text{NLL} = -\sum_{x\in \mathcal{D}} log(P(y|x))},
    \mathcal{L}_{total} = \lambda_1 NLL + (1-\lambda_1) \mathcal{L}_{alignment}
\end{equation}
where $y$ is the true value, $x$ is the input data, $\mathcal{D}$ is the training set, $P(y|x)$ is the predicted conditional probability of $y$ given $x$, and $\lambda_1$ is the loss component hyperparameter. In multivariate modeling, $NLL$ can be modeled differently. In Strada-LLM, we estimate the joint distribution, but further experiments are discussed in the ablation study~\ref{nll_ablation}. 

% In contrast to other loss functions, such as Mean Absolute Error (MAE), Root Mean Squared Error (RMSE), and Mean Absolute Percentage Error (MAPE)~\cite{li2024flashstsimpleuniversalprompttuning}, which measure the difference between the predicted and true values directly, the negative log-likelihood measures the conditional probability of the true value under the model's predictions.

\section{EXPERIMENTS}\label{exp}

\begin{figure}
\centering
  \includegraphics[width=0.40\textwidth,keepaspectratio]{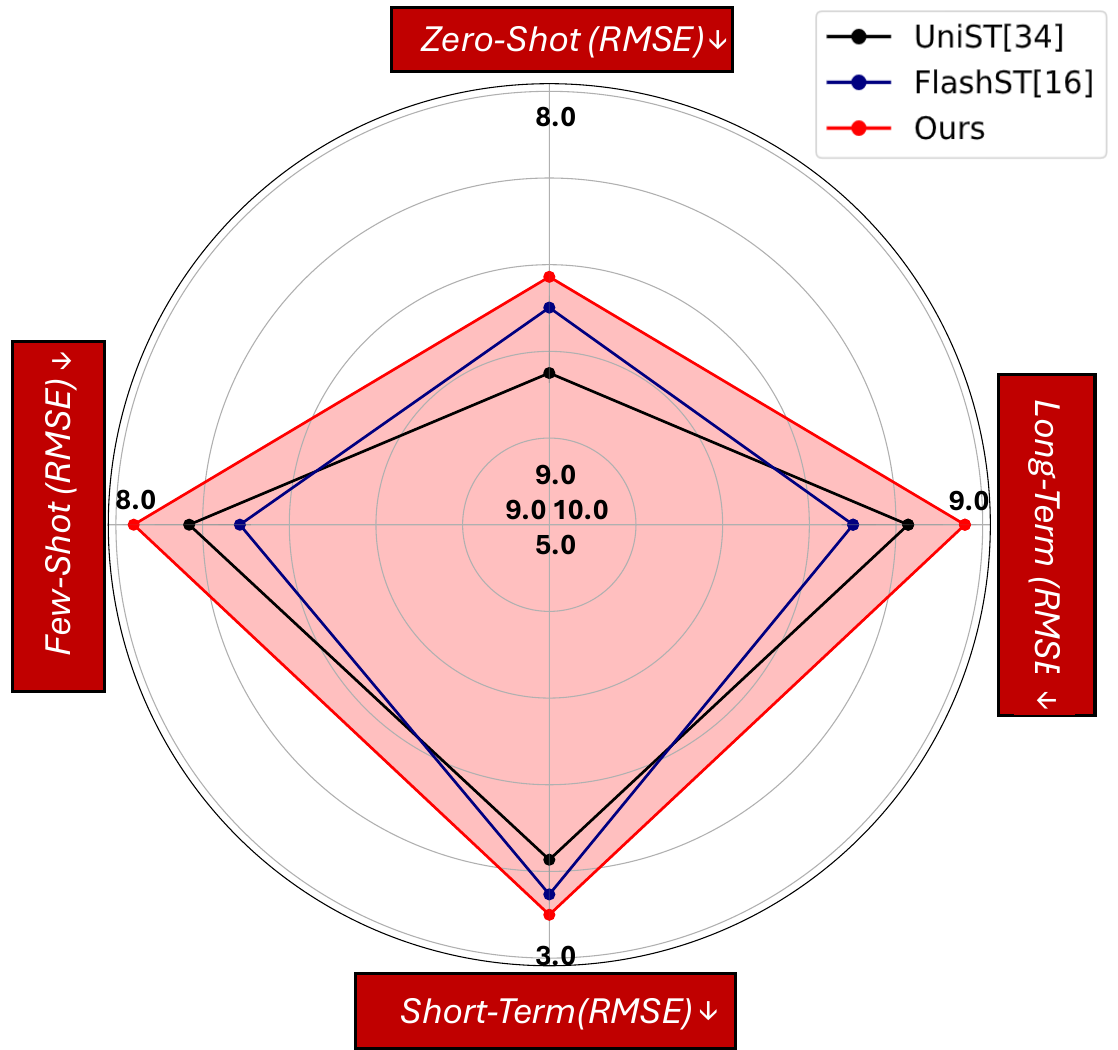}
  \caption{Comparison of unified benchmark for different LLMs.}
  \label{fig:teaser}
\end{figure}

\subsection{Datasets Details}

In this section, we assess the Strada-LLM model's predictive capabilities using eight real-world datasets. The description of these datasets is presented in Table~\ref{tab:dataset_description}. Most of these datasets pertain to traffic speed. For the sake of simplicity, traffic speed is utilized as the primary traffic metric in our experimental analysis; thus one can presume $F=1$. {Detailed training strategies, e.g., hyperparameters and experimental setup, are explained in Appendix~\ref{strategy}}.

\begin{table}[ht]
    \centering
    \caption{Dataset Description}
    \label{tab:dataset_description}
    \begin{adjustbox}{width=0.48\textwidth}
    \begin{tabular}{@{}llccc@{}}
        \hline
        \textbf{Dataset} & \textbf{Data Type} & \textbf{Number of Nodes} & \textbf{Region} & \textbf{Time Steps} \\
        \hline
        \multicolumn{4}{l}{{Pretraining Datasets}} \\
        \hline
        PeMS03~\cite{10.1145/3447548.3467430}          & Traffic Speed      & 170 & California & 26208 \\
        PeMS04~\cite{10.1145/3447548.3467430}          & Traffic Speed      & 307 & California & 16992 \\
        PeMS07~\cite{10.1145/3447548.3467430}          & Traffic Speed      & 883 & California & 26208 \\
        PeMS08~\cite{10.1145/3447548.3467430}          & Traffic Flow      & 170 & California & 17856 \\
        \hline
        \multicolumn{4}{l}{{Few-Shot}} \\
        \hline
        METR-LA~\cite{10.1145/3447548.3467430}         & Traffic Speed      & 207 & California & 34272 \\
        PEMS-Bay~\cite{10.1145/3447548.3467430}        & Traffic Speed      & 325 & California & 52116 \\
        PeMS07(M)~\cite{10.1145/3447548.3467430}       & Traffic Speed      & 228 & California & 12672 \\
        Brussels~\cite{10591664}        & Traffic Count      & 207 & Brussels & 12672 \\
        \hline
                \multicolumn{4}{l}{{Zero-Shot}} \\
        \hline
        Crowd~\cite{unist}       & 
        Pedestrian Count & $16 \times 20$ & Nanjing & 9420 \\
        PeMS07(M)~\cite{10.1145/3447548.3467430}        & Traffic Count      & 228 & California & 12672 \\
        Brussels~\cite{10591664}        & Traffic Count      & 207 & Brussels & 12672 \\
        \hline
    \end{tabular}
\end{adjustbox}
\end{table}

\subsection{Baselines}
We compare the performance of the Strada-LLM model with \textbf{spatio-temporal} methods, GMAN ~\cite{Zheng_Fan_Wang_Qi_2020}, MTGNN ~\cite{10.1145/3394486.3403118}, GTS ~\cite{shang2021discrete}, STEP ~\cite{10.1145/3534678.3539396}, and \textbf{LLM-based} methods which either are \textbf{point-wise forecasters} such as FlashST \cite{li2024flashstsimpleuniversalprompttuning}, UniST \cite{unist} or \textbf{probabilistic forecasters} such as Lag-llama\cite{rasul2024lagllamafoundationmodelsprobabilistic}. The details of the baseline methods are explained in Appendix \ref{app_base}. To compare with the baselines, we adopt the MAE, RMSE, and MAPE prediction error metrics~\cite{li2024flashstsimpleuniversalprompttuning}. Due to Strada-LLM being a probabilistic model, we also compare with Lag-Llama~\cite{rasul2024lagllamafoundationmodelsprobabilistic} using the Continuous Ranked Probability Score (CRPS) metric, defined as:
$$
\small{
CRPS(F, y) = \int_{-\infty}^{\infty} (F(x) - H(x - y))^2 dx,
}
$$
where $F(x)$ is the cumulative distribution function (CDF) of the forecasted distribution, $y$ is the observed value, and $H(x - y)$ is the Heaviside step function, which is 0 when $x < y$ and 1 when $x \geq y$.

\subsection{{LLM Forecasting Results}}\label{f_result}

\subsubsection{{LLM Short-Term Forecasting}}\label{f_short_result}
The short-term traffic prediction accuracy was evaluated across four datasets: PeMS-Bay, METR-LA, PEMS07(M), and Brussels. The last dataset is private. Strada-LLM, as an LLM should be benchmarked jointly for all datasets. 
% So, we report the performance in {two manners}:
% \begin{enumerate}
%     \item \textbf{knowledge adaptation}: Like every LLM, we pre-trained Strada-LLM on the datasets PeMS03, PeMS04, PeMS07, and PeMS08 then fine tuned on the PeMS-Bay, METR-LA, \\PEMS07(M), and Brussels datasets in few-shot fashion; we refer to this approach as \textit{LLM} setup.
%     \item  \textbf{fully-supervised}: We trained our proposed model on a single dataset in a fully-supervised fashion (referred to as \textit{$Strada-LLM\left[Solo\right]$} in this section).
% \end{enumerate}

The prediction results for the datasets METR-LA and PEMS-Bay are reported in Tables~\ref{tab1}--\ref{tab2}, respectively. 
The ability to perform forecasting and being robust to new domains are important qualities for an LLM. As shown in Tables \ref{tab1}-- \ref{tab2}, the proposed LLM has competitive performance compared to STEP~\cite{10.1145/3534678.3539396}, which is the corresponding transformer-based fully-supervised method. Readers can refer to Appendix~\ref{fig4p} about robustness analysis. On the other hand, compared to the best prompt-tuning method FlashST~\cite{li2024flashstsimpleuniversalprompttuning}, shows that taking the global structure of the graph into account brings forth a definite advantage. Moreover, although Strada-LLM and STEP~\cite{10.1145/3534678.3539396} are both transformer-based models, the introduction of normalization blocks like RMSNorm~\cite{Zhou_Zhang_Peng_Zhang_Li_Xiong_Zhang_2021} and Rotary Positional Encoding (RoPE) differentiates them. Additionally, while STEP~\cite{10.1145/3534678.3539396} is an encoder-decoder model, Strada-LLM is a decoder-only model, which makes it optimized in terms of performance for inference~\cite{fu2023decoderonlyencoderdecoderinterpretinglanguage}. As we can see in Tables~\ref{tab1}--\ref{tab2}, the proposed Strada-LLM manages to outperform the baseline models on the aforementioned datasets in the longer prediction horizon more significantly. Specifically, on the PeMS-Bay dataset, our proposed model achieves an RMSE of 3.94 for 1-hour prediction, corresponding to an improvement of approximately $16\%$ compared to the prompt-tuning baseline~\cite{li2024flashstsimpleuniversalprompttuning}.

\begin{table*}[htbp]
% \begin{threeparttable}
\caption{{Traffic short-term forecasting comparison with the state-of-the-art methods on the METR-LA dataset}}
\begin{center}
\begin{tabular}{@{}l@{\hspace{3mm}}l|ccc|ccc|ccc@{}}
\toprule
 & \multirow{2}{*}{Baselines} & \multicolumn{3}{c|}{15 min} & \multicolumn{3}{c|}{30 min} & \multicolumn{3}{c}{60 min} \\ \cmidrule{3-11}
& & MAE$\downarrow$ & RMSE$\downarrow$ & MAPE(\%)$\downarrow$ & MAE$\downarrow$ & RMSE$\downarrow$ & MAPE(\%)$\downarrow$ & MAE$\downarrow$ & RMSE$\downarrow$ & MAPE(\%)$\downarrow$ \\

\midrule
\multirow{4}{*}{\rotatebox[origin=c]{90}{Supervised}} 
 & GMAN \cite{Zheng_Fan_Wang_Qi_2020} & 2.80 & 5.55 & 7.41 & 3.12 & 6.49 & 8.73 & 3.44 & 7.35 & 10.07 \\ \cline{2-11}
 & MTGNN \cite{10.1145/3394486.3403118} &  2.69 & 5.18 & 6.88 & 3.05 & 6.17 & 8.19 & 3.49 & 7.23 & 9.87 \\ \cline{2-11}
 & GTS \cite{shang2021discrete} & 2.67 & 5.27 & 7.21 & 3.04 & 6.25 & 8.41 & 3.46 & 7.31 & 9.98 \\ \cline{2-11}
 & STEP\cite{10.1145/3534678.3539396} & \underline{2.61} & \underline{4.98} & \underline{6.6} & \underline{2.99} & \underline{5.97} & \underline{7.96} & \underline{3.37} & \underline{6.99} & \underline{9.61} \\
\midrule
\multirow{3}{*}{\rotatebox[origin=c]{90}{LLM}} 
 & FlashST \cite{li2024flashstsimpleuniversalprompttuning} & {2.84} & {5.57} & {7.45} & {3.19} & {6.43} & {9.04} & {3.60} & {7.44} & {10.68}  \\ \cline{2-11}
 & UniST \cite{unist} & {2.89} & {5.63} & {7.49} & {3.26} & {6.51} & {9.09} & {3.65} & {7.52} & {10.79}  \\ \cline{2-11}
 & Strada-LLM({ours}) & \textbf{2.45} & \textbf{4.71} & \textbf{6.42} & \textbf{2.97} & \textbf{5.84} & \textbf{7.87} & \textbf{3.35} & \textbf{6.73} & \textbf{9.47} \\
\bottomrule
\end{tabular}
\label{tab1}
\end{center}
\end{table*}

\begin{table*}[htbp]
\caption{{Traffic short-term predictions comparison with the state-of-the-art methods on PEMS-Bay dataset}}
\begin{center}
\begin{tabular}{@{}l@{\hspace{3mm}}l|ccc|ccc|ccc@{}}
\toprule
 & \multirow{2}{*}{Baselines} & \multicolumn{3}{c|}{15 min} & \multicolumn{3}{c|}{30 min} & \multicolumn{3}{c}{60 min} \\ \cmidrule{3-11}
& & MAE$\downarrow$ & RMSE$\downarrow$ & MAPE(\%)$\downarrow$ & MAE$\downarrow$ & RMSE$\downarrow$ & MAPE(\%)$\downarrow$ & MAE$\downarrow$ & RMSE$\downarrow$ & MAPE(\%)$\downarrow$ \\
\midrule
\multirow{4}{*}{\rotatebox[origin=c]{90}{Supervised}} 
& GMAN \cite{Zheng_Fan_Wang_Qi_2020} & 1.34 & 2.91 & 2.86 & 1.63 & 3.76 & 3.68 & 1.86 & 4.32 & 4.37 \\ \cline{2-11}
& MTGNN \cite{10.1145/3394486.3403118} & 1.32 & 2.79 & 2.77 & 1.65 & 3.74 & 3.69 & 1.94 & 4.49 & 4.53 \\ \cline{2-11}
& GTS \cite{shang2021discrete} & 1.34 & 2.83 & 2.82 & 1.66 & 3.78 & 3.77 & 1.95 & 4.43 & 4.58  \\ \cline{2-11}
& STEP\cite{10.1145/3534678.3539396} & \textbf{1.26} & \textbf{2.73} & \textbf{2.59} & \underline{1.55} & \underline{3.58} & \underline{3.43} & \underline{1.79} & \underline{4.20} & \underline{4.18} \\
\midrule
\multirow{3}{*}{\rotatebox[origin=c]{90}{LLM}}
& FlashST \cite{li2024flashstsimpleuniversalprompttuning}  & {1.37} & {2.96} & {2.88} & {1.72} & {3.98} & {3.89} & {2.02} & {4.69} & {4.79}  \\ \cline{2-11}
& UniST \cite{unist} & {1.42} & {3.01} & {2.91} & {1.78} & {3.99} & {3.91} & {2.07} & {4.72} & {4.81}  \\ \cline{2-11}
& Strada-LLM ({Ours})  & \underline{1.28} & \underline{2.77} & \underline{2.61} & \textbf{1.53} & \textbf{3.14} & \textbf{3.37} & \textbf{1.76} & \textbf{3.94} & \textbf{4.09} \\
\bottomrule
\end{tabular}
\label{tab2}
\end{center}
\end{table*}

\begin{table*}[htbp]
\centering
\caption{{Long-Term prediction experiment on the PEMS07 (M) and Brussels dataset ($p\text{-value}<0.05$)}}
\begin{center}
\begin{adjustbox}{width=1\textwidth}
\begin{tabular}{l|ccc|ccc|ccc|ccc|ccc|ccc}
\toprule
& \multicolumn{9}{c|}{PEMS07(M)} & \multicolumn{9}{c}{Brussels}\\
\cmidrule{2-19}
\multirow{2}{*}{Baselines} & \multicolumn{3}{c|}{90 min} & \multicolumn{3}{c|}{120 min} & \multicolumn{3}{c|}{150 min} & \multicolumn{3}{c|}{90 min} & \multicolumn{3}{c|}{120 min} & \multicolumn{3}{c}{150 min} \\ \cmidrule{2-19}
& MAE$\downarrow$ & RMSE$\downarrow$ & MAPE(\%)$\downarrow$ & MAE$\downarrow$ & RMSE$\downarrow$ & MAPE(\%)$\downarrow$ & MAE$\downarrow$ & RMSE$\downarrow$ & MAPE(\%)$\downarrow$ & MAE$\downarrow$ & RMSE$\downarrow$ & MAPE(\%)$\downarrow$ & MAE$\downarrow$ & RMSE$\downarrow$ & MAPE(\%)$\downarrow$ & MAE$\downarrow$ & RMSE$\downarrow$ & MAPE(\%)$\downarrow$ \\
\midrule
FlashST \cite{li2024flashstsimpleuniversalprompttuning} & 4.21 & 6.98 & \underline{10.71} & 4.51 & 7.20 & 11.92 & 4.83 & 7.54 & 11.38 & 6.35 & 8.53 & 9.57 & 6.47 & 8.89 & 9.84 & \underline{6.79} & \underline{9.31} & 10.18 \\ \hline
UniST \cite{unist} & \underline{3.98} & \underline{6.71} & 10.84 & \underline{4.09} & \underline{6.66} & \underline{11.02} & \underline{4.34} & \underline{7.47} & \underline{11.25} & \underline{6.27} & \underline{8.51} & \underline{9.54} & \underline{6.42} & \underline{8.85} & \underline{9.82} & {6.82} & 9.32 & \underline{10.14}\\ \hline
Strada-LLM(ours) & \textbf{3.84} & \textbf{6.02} & \textbf{10.53} & \textbf{4.04} & \textbf{6.51} & \textbf{11.01} & \textbf{4.19} & \textbf{7.01} & \textbf{11.08} & \textbf{5.83} & \textbf{8.31} & \textbf{9.39} & \textbf{6.21} & \textbf{8.70} & \textbf{9.77} & \textbf{6.55} & \textbf{9.01} & \textbf{10.03}\\
\bottomrule
\end{tabular}
\end{adjustbox}
\label{tab_log_term}
\end{center}
\end{table*}

\subsubsection{{LLM Long-Term Forecasting}}\label{f_long_result}
Followed by the setup of ~\cite{unist}, we experiment with long-term forecasting for horizons \textbf{90, 120, and 150} minutes. The long-term traffic prediction accuracy was evaluated across two datasets: Crowd and PEMS07(M). We report the performance in Table~\ref{tab_log_term}. Consistently, Strada-LLM outperforms the LLM-based baselines. The successful performance can be attributed to the Graphical modeling of spatial covariates. Further visualizations are presented in the Appendix~\ref{vis}.

\begin{table*}[htbp]
\centering
\caption{{Few-shot learning experiment on the PEMS07(M) and Brussels dataset($p\text{-value}<0.05$)}}
\begin{center}
\begin{adjustbox}{width=1\textwidth}
\begin{tabular}{l|ccc|ccc|ccc|ccc|ccc|ccc}
\toprule
& \multicolumn{9}{c|}{PEMS07(M)} & \multicolumn{9}{c}{Brussels}\\
\cmidrule{2-19}
\multirow{2}{*}{Baselines} & \multicolumn{3}{c|}{15 min} & \multicolumn{3}{c|}{30 min} & \multicolumn{3}{c|}{60 min} & \multicolumn{3}{c|}{15 min} & \multicolumn{3}{c|}{30 min} & \multicolumn{3}{c}{60 min} \\ \cmidrule{2-19}
& MAE$\downarrow$ & RMSE$\downarrow$ & MAPE(\%)$\downarrow$ & MAE$\downarrow$ & RMSE$\downarrow$ & MAPE(\%)$\downarrow$ & MAE$\downarrow$ & RMSE$\downarrow$ & MAPE(\%)$\downarrow$ & MAE$\downarrow$ & RMSE$\downarrow$ & MAPE(\%)$\downarrow$ & MAE$\downarrow$ & RMSE$\downarrow$ & MAPE(\%)$\downarrow$ & MAE$\downarrow$ & RMSE$\downarrow$ & MAPE(\%)$\downarrow$ \\
\midrule
FlashST \cite{li2024flashstsimpleuniversalprompttuning} & {3.37} & 5.14 & 5.93 & {3.59} & 5.25 & 6.84 & {3.83} & 6.81 & \underline{7.46} & 4.73 & 7.54 & {8.34} & \underline{5.24} & 8.59 & {9.53} & 6.06 & 9.01 & {10.04} \\ \hline
UniST \cite{unist} &  \underline{2.45} & \underline{4.89} & \underline{5.06} & \underline{2.67} & \underline{5.13} & \underline{6.56} & \underline{2.89} & \underline{6.59} & 7.50 & \underline{4.70} & \underline{7.29} & \underline{8.09} & 5.29 & \underline{8.42} & \underline{9.34} & \underline{5.90} & \underline{8.96} & \underline{9.94} \\ \hline
Strada-LLM(ours) & \textbf{2.38} & \textbf{4.13} & \textbf{4.92} & \textbf{2.60} & \textbf{4.86} & \textbf{6.51} & \textbf{2.84} & \textbf{6.43} & \textbf{7.41} & \textbf{4.53} & \textbf{7.16} & \textbf{8.05} & \textbf{5.16} & \textbf{8.14} & \textbf{9.03} & \textbf{5.86} & \textbf{8.57} & \textbf{9.52}\\
\bottomrule
\end{tabular}
\end{adjustbox}
\label{tab1.5}
\end{center}
\end{table*}

\subsubsection{{LLM Zero-Shot Forecasting}}\label{f_mvf_result_zero}
The zero-shot traffic prediction accuracy was evaluated across three datasets: Crowd, Pems07(M), and Brussels. Table ~\ref{tab_zero_shot} presents the results of zero-shot capability of Strada-LLM compared to the other LLMs. %Spatial correlation capturing contribution causes performance enhancement.
Further visualizations are presented in the Appendix~\ref{vis}. Strada-LLM is able to capture the spatio-temporal dependencies between connected nodes and prediction correlated trends, which is neglected by UniST~\cite{unist}. To demonstrate the generalizability to the other urban datasets, Strada-LLM's performance on the Crowd dataset~\cite{unist} is demonstrated in Figure~\ref{crowd_zero}, which outperforms other LLM baselines.

\begin{figure}
\centering
\begin{tikzpicture}
\begin{axis}[
    width=8cm,
    height=5cm,
    ybar,
    bar width=11pt,
    ylabel={MAE},
    xlabel={Horizons},
    title={},
    symbolic x coords={15min,30min,60min},
    xtick=data,
    ymin=6.2,
    ymax=8.0,
    legend style={
        at={(axis description cs:-0.10,-0.15)},
        anchor=north west,
        legend columns=3,
        column sep=0.5cm,  % Uniform spacing between legend entries
        font=\small,
        /tikz/every even column/.append style={column sep=0.5cm},  % Make spacing consistent
        /tikz/every odd column/.append style={column sep=0.5cm},
        draw
    },
    legend image code/.code={
        \draw[#1] (0cm,-0.1cm) rectangle (0.4cm,0.1cm);
    },
    legend cell align={left},
    nodes near coords,
    nodes near coords style={font=\small},
    % Increase font sizes
    tick label style={font=\small},
    label style={font=\small},
    title style={font=\small},
    legend style={font=\small}
]

% Model A data with forward diagonal pattern
\addplot[draw=blue,fill=blue!30,pattern=north east lines] 
    coordinates {(15min,6.53) (30min,6.87) (60min,7.17)};

% Model B data with backward diagonal pattern
\addplot[draw=orange,fill=orange!30,pattern=north west lines] 
    coordinates {(15min,6.86) (30min,7.05) (60min,7.31)};

% Model C data with crosshatch pattern
\addplot[draw=green,fill=green!30,pattern=crosshatch] 
    coordinates {(15min,7.02) (30min,7.12) (60min,7.63)};

\legend{\textbf{Ours}, UniST\cite{unist}, FlashST\cite{li2024flashstsimpleuniversalprompttuning}}
\end{axis}
\end{tikzpicture}
\caption{Zero-shot performance comparison on Crowd dataset.}
\label{crowd_zero}
\end{figure}

\subsubsection{{LLM Probabilistic Forecasting}}\label{f_mvf_result}
Table \ref{tab:probab} reports the CRPS metric, likewise Lag-llama\cite{rasul2024lagllamafoundationmodelsprobabilistic}, it has been implemented as the mean quantile loss. We
observe that Strada-LLM consistently
achieve strong few-shot and few-shot performance, obtaining either the best
or second-best results for the Pems07 and Brussels datasets.

\begin{table}[htbp]
\centering
\caption{{Probabilistic forecasting experiment on the PEMS07(M) and Brussels dataset}}
\resizebox{\columnwidth}{!}{%
\begin{tabular}{lccccccc}
\toprule
& \multicolumn{3}{c}{PEMS07(M)} & \multicolumn{3}{c}{Brussels} \\
\cmidrule(lr){2-4} \cmidrule(lr){5-7}
Baselines &  \multicolumn{3}{c}{CRPS$\downarrow$} & \multicolumn{3}{c}{CRPS$\downarrow$} \\
& 15 min & 30 min & 60 min & 15 min & 30 min & 60 min \\
\midrule
Lag-llama~\cite{rasul2024lagllamafoundationmodelsprobabilistic} & \underline{3.95} & \underline{4.13} & \underline{4.92} & \underline{5.07} & \underline{5.77} & \underline{5.91} \\
Strada-LLM(ours) & \textbf{3.13} & \textbf{3.67} & \textbf{4.03} & \textbf{4.93} & \textbf{5.10} & \textbf{5.31} \\
\bottomrule
\end{tabular}
}
\label{tab:probab}
\end{table}
\begin{table*}[htbp]
\centering
\caption{{Zero-shot learning experiment on the PEMS07(M) and Brussels dataset($p\text{-value}<0.05$)}}
\begin{center}
\begin{adjustbox}{width=1\textwidth}
\begin{tabular}{l|ccc|ccc|ccc|ccc|ccc|ccc}
\toprule
& \multicolumn{9}{c|}{PEMS07(M)} & \multicolumn{9}{c}{Brussels}\\
\cmidrule{2-19}
\multirow{2}{*}{Baselines} & \multicolumn{3}{c|}{15 min} & \multicolumn{3}{c|}{30 min} & \multicolumn{3}{c|}{60 min} & \multicolumn{3}{c|}{15 min} & \multicolumn{3}{c|}{30 min} & \multicolumn{3}{c}{60 min} \\ \cmidrule{2-19}
& MAE$\downarrow$ & RMSE$\downarrow$ & MAPE(\%)$\downarrow$ & MAE$\downarrow$ & RMSE$\downarrow$ & MAPE(\%)$\downarrow$ & MAE$\downarrow$ & RMSE$\downarrow$ & MAPE(\%)$\downarrow$ & MAE$\downarrow$ & RMSE$\downarrow$ & MAPE(\%)$\downarrow$ & MAE$\downarrow$ & RMSE$\downarrow$ & MAPE(\%)$\downarrow$ & MAE$\downarrow$ & RMSE$\downarrow$ & MAPE(\%)$\downarrow$ \\
\midrule
FlashST \cite{li2024flashstsimpleuniversalprompttuning} & \underline{2.77} & \underline{5.32} & \underline{6.01} & \textbf{2.91} & \underline{5.61} & \textbf{6.75} & \underline{3.11} & 7.11 & 7.98 & 4.94 & 7.92 & {8.79} & \underline{5.89} & \underline{8.86} & \underline{9.83} & \underline{6.46} & \underline{9.42} & \underline{10.45} \\ \hline
UniST \cite{unist} & 2.92 & {5.42} & {6.05} & 2.99 & \textbf{5.49} & 6.92 & 3.31 & \underline{7.02} & \underline{7.90} & \underline{4.92} & \underline{7.69} & \underline{8.53} & 5.92 & 8.88 & 9.85 & 6.49 & 9.59 & 10.64 \\ \hline
Strada-LLM(ours) & \textbf{2.76} & \textbf{4.77} & \textbf{5.84} & \underline{2.93} & {5.80} & \underline{6.79} & \textbf{3.09} & \textbf{6.81} & \textbf{8.10} & \textbf{4.90} & \textbf{7.48} & \textbf{ 8.30} & \textbf{5.86} & \textbf{8.73} & \textbf{9.69} & \textbf{6.36} & \textbf{9.39} & \textbf{10.42}\\
\bottomrule
\end{tabular}
\end{adjustbox}
\label{tab_zero_shot}
\end{center}
\end{table*}

\subsection{Ablation Study}\label{ablation}

\begin{table*}[htbp]
\caption{{Domain adaptation comparison for demonstrating selection of effective source dataset}}
\begin{center}
\begin{tabular}{lc|cc|cc|cc|cc}
\toprule
\multicolumn{2}{c|}{Source $\rightarrow$ Target}\ & \multicolumn{2}{c|}{$Top^k$ Fine-tuning} & \multicolumn{2}{c|}{\textbf{Ours}} & \multicolumn{2}{c|}{Fully-Finetune} & \multicolumn{2}{c}{Fully-Supervised} \\ \cmidrule{3-10}
&  & MAE$\downarrow$ & RMSE$\downarrow$ & MAE$\downarrow$ & RMSE$\downarrow$ & MAE$\downarrow$ & RMSE$\downarrow$ & MAE$\downarrow$ & RMSE$\downarrow$ \\
\midrule
Metr-LA & Pems-BAY & 3.31 & 5.76 & 2.64 & 4.83 & 2.01 & 3.89 & 1.54 & 3.53 \\ \hline
% Metr-LA & PeMS04 & 22.36 & 28.72 & 22.12 & 27.58 & 20.41 & 26.43 & 19.26 & 25.47 \\
PeMS04 & Pems-BAY & 2.73 & 5.02 & 1.96 & 4.15 & 1.67 & 3.67 & 1.54 & 3.53 \\
\bottomrule
\end{tabular}
\label{tab4}
\end{center}
\end{table*}

\begin{table}
\centering
\caption{The computation cost on the PEMS07(M) with the batchsize=64.}
\begin{tabular}{@{}lccc@{}}
\hline
Method & Parameters(M) & \makecell{Trainable\\Parameters(M)} & \makecell{Trainable\\Ratio(\%)$\downarrow$} \\
\hline
FlashST\cite{li2024flashstsimpleuniversalprompttuning} & 64.02 & 1.01 & \underline{1.58} \\
UniST\cite{unist} & 30.00 & 0.71 & 2.39 \\
Strada-LLM (ours) & 40.00 & 0.52 & \textbf{1.32} \\
\hline
\end{tabular}
\label{tab_comp}
\end{table}

\subsubsection{{Domain Adaptation}}
We examine the adaptation capabilities of Strada-LLM by evaluating the gap when different datasets are used as source and target domains. Specifically, we trained Strada-LLM on one dataset (e.g., PeMS-04) and adapted to another (e.g., METR-LA) to assess its generalization across different types of traffic patterns. Table~\ref{tab4} presents the results for the domain adaptation process when pre-trained on a source dataset and adapted to the corresponding target dataset. The results presented in Table~\ref{tab4} indicate that there is a performance drop when the model is applied to a different target domain. This table can also address selecting efficient pre-training datasets to gain transferability advance for the targeted LLM. Take Pems-Bay as an example, selecting Pems04 as the pertaining dataset for LLM ends up the 14.07\% superior performance in the adaptation phase compared to Metr-LA, because of the inherent closer distribution. 
% Therefore, considering \textbf{Pems-bay and Metr-LA} as the pertaining datasets will have more distribution coverage compared to Metr-LA and PeMS04. 
{Detailed ablated studies about the performance of LoRA in terms of the rank parameter are discussed in Appendix~\ref{rank_analyse}.}

% , Strada-LLM still maintains a competitive accuracy compared to baseline models. This performance can be attributed to Strada-LLM's robust architecture and its ability to learn transferable features across datasets. For instance, when trained on PeMS-Bay and tested on METR-LA, Strada-LLM achieved a MAPE of 12.34\%, which is only a slight increase from its performance within the same domain. This demonstrates the model's potential in real-world scenarios where traffic data from different cities need to be analyzed without extensive retraining, highlighting the importance of domain adaptation in creating versatile and scalable traffic prediction models.  

\begin{figure}[!htb]
\includegraphics[width=0.30\textwidth, keepaspectratio]{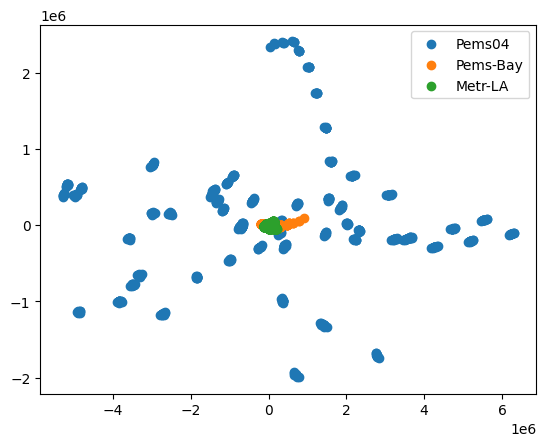}
\centering\caption{{Latent space embedding of the last layer (before the distribution head) visualized in a 2D space.}}
% \centering\caption{proposed LoRA method for transfer learning}
\label{latent}
\end{figure}

\subsubsection{{Model Interpretation}}
To better understand Strada-LLM, we visualize the latent space of the LLM representation for different datasets. By doing so, the disparity of dataset distributions and the reason for the difficulty of adapting to PeMS04 are revealed. {The visualization is shown in Figure~\ref{latent} which is obtained with T-SNE~\cite{cai2022theoreticalfoundationstsnevisualizing} in 2D dimensions.} Our variational posterior distribution $P(\theta \mid X)$ can truly depict the underlying distribution.  

% \begin{figure}[htb]

% \includegraphics[width=\textwidth]{fig2.pdf}
% \subfigure[Latent space embedding of last layer(before the distribution head) visualization in 2D space]{
%         \includegraphics[width=0.45\textwidth]{2d_embedding.png}
%     }
%     \hfill
%     \subfigure[visualization in 3D space of the latent space embedding of the last layer, taking place before the distribution head.]{
%         \includegraphics[width=0.45\textwidth]{3d_embedding.png}
%     }
% \centering\caption{Domain adaptation result for Strada-LLM by applying different approaches}
% % \centering\caption{proposed LoRA method for transfer learning}
% \label{latent}
% \end{figure}

\subsubsection{Scalability of K-hop structure}\label{scalability_section}
{
% We evaluate the performance of our Strada-LLM with different hop's radius. To do so, we try various values for $k \in (1, 3, 5)$, and the results are demonstrated in Table~\ref{tab_khop}. As can be seen, the lower error comes with the cost of increasing the neighboring radius. As shown in Table~\ref{tab:probab}, Strada-LLM outperforms Lag-Llama~\cite{rasul2024lagllamafoundationmodelsprobabilistic} which did not take spatial dependencies into account while our model incorporates those by utilizing K-hop block.
Our experimental analysis explores Strada-LLM's sensitivity to different neighborhood radii by examining performance across multiple k-hop values ($k \in \{1, 3, 5\}$). The results in Table~\ref{tab_khop} demonstrate a trade-off between prediction accuracy and neighborhood size, with larger radii yielding lower error rates. Comparative analysis presented in Table~\ref{tab:probab} shows that Strada-LLM achieves superior performance over Lag-Llama~\cite{rasul2024lagllamafoundationmodelsprobabilistic}, which did not take neighbors into account, primarily due to our model's incorporation of spatial dependencies through the K-hop block structure, a feature absent in the baseline approach.
}

\begin{table}[htbp]
\caption{Feature Scalability Analysis Benchmark}
\begin{center}
\begin{tabular}{lccc}
\toprule
$K$ & MAE$\downarrow$ & RMSE$\downarrow$ & MAPE(\%)$\downarrow$ \\
\midrule
$1$ & 2.96 & 5.34  & 7.35 \\
$3$ & \textbf{2.60} & \textbf{4.86} & \textbf{6.51} \\
$5$ & \underline{2.63} & \underline{4.89}  & \underline{6.54} \\
\bottomrule
\end{tabular}
\label{tab_khop}
\end{center}
\end{table}

\subsubsection{Computation Cost}\label{sec_comp}

{We evaluate the performance of our proposed LoRA-based approach compared to existing LLMs. Results are demonstrated in Table~\ref{tab_comp}. As can be seen, the lower error comes with the cost of involving more parameters. However, in contrast to UniST~\cite{unist}, Strada-LLM's adaptation outperforms the compression rate, from 1.58(\%) to 1.32(\%), which demonstrates its efficiency. For real-time applications, more lightweight models are being demanded, which remains the future direction. Performance over parameter analysis is illustrated in Appendix~\ref{scalability}.}

\subsubsection{{Distribution modeling}}\label{nll_ablation}

{We evaluate our model's flexibility by deploying it across multiple LLM architectures. So, we report the performance in Table~\ref{tab_dist}. Consistently, Strada-LLM outperforms the LLM-based baselines. The successful performance can be attributed to the Graphical modeling of spatial covariates. }

\begin{table}[htbp]
\caption{Ablation study on distribution modeling for PEMS07(M)}
\begin{center}
\begin{tabular}{lcccc}
\toprule

Backbone & MAE$\downarrow$ & RMSE$\downarrow$ & MAPE(\%)$\downarrow$\\
\midrule
Univariate & 2.81 & 5.63 & 6.81 \\
Independent & \underline{2.69} & \underline{5.37} & \underline{6.73} \\
Joint & \textbf{2.60} & \textbf{4.86} & \textbf{6.51} \\
\bottomrule
\end{tabular}
\label{tab_dist}
\end{center}
\end{table}

\subsubsection{{LLM Cross-Compatibility}}

We evaluate our model's flexibility by deploying it across multiple LLM architectures. Specifically, we utilized Mistral and Llama 2 as base models to assess the adaptability of our proposed method. The results are shown in Table \ref{tab6}.

\begin{table}[htbp]
\caption{Ablation study on LLM backbone adaptability for PEMS07(M)}
\begin{center}
\begin{tabular}{lcccc}
\toprule

Backbone & MAE$\downarrow$ & RMSE$\downarrow$ & MAPE(\%)$\downarrow$\\
\midrule
Llama2~\cite{rasul2024lagllamafoundationmodelsprobabilistic}  & 2.83 & 5.66 & 6.89 \\
Mistral~\cite{jiang2023mistral7b} & \textbf{2.60} & \textbf{4.86} & \textbf{6.51} \\
\bottomrule
\end{tabular}
\label{tab6}
\end{center}
\end{table}

\section{Conclusion}

This research proposes a probabilistic-based LLM for traffic forecasting called Strada-LLM, which injects the local graph structure implicitly into the input tokens. We utilize a subgraph extraction procedure to inject the graph structure into the transformer token inputs. On the one
hand, we adopt a probabilistic transformer to predict the traffic data by sampling from the underlying distribution.
On the other hand, we utilize a low-rank method for the transfer learning task. Consequently, we showcase its success in adapting to datasets with significant differences compared to the datasets used for pre-training. Finally, Strada-LLM is compliant to a variety of known LLM backbones. In summary, the Strada-LLM model successfully captures
the spatial and temporal features from traffic data so that they can
be applied to other spatio-temporal tasks. A potential avenue for future research could involve evaluating the expressiveness of the distribution head.
% , interpretable, and scalable

% \begin{acks}
% This work is funded by Innoviris within the research project TORRES. The authors thank Leandro Di Bella, Weijiang Xiong, and Bert Van Hauwermeiren for comments that greatly improved the manuscript.

% \end{acks}

\bibliographystyle{ACM-Reference-Format}
\bibliography{bibfile}

% \pagebreak
\appendix

\section{Appendix}

\subsection{Baselines}\label{app_base}

We compare the performance of the Strada-LLM model with the following {baseline methods}:
\begin{itemize}
    \item GMAN ~\cite{Zheng_Fan_Wang_Qi_2020}: The gating fusion mechanism redefines the spatio-temporal attention block.
    \item MTGNN ~\cite{10.1145/3394486.3403118}: Alternating use of graph convolution and temporal convolution modules.
    \item GTS ~\cite{shang2021discrete}: The underlying graph structure is learned among multiple time series, and the corresponding time series is simultaneously predicted with DCRNN~\cite{li2018diffusion}.
    \item STEP ~\cite{10.1145/3534678.3539396}: Adopts the graph wavenet model~\cite{ijcai2019p264}, integrated into the transformer backbone and pre-training scheme.
    \item FlashST \cite{li2024flashstsimpleuniversalprompttuning}, UniST \cite{unist}: The data distribution shift is learned by leveraging prompt-tuning.
    \item Lag-Llama~\cite{rasul2024lagllamafoundationmodelsprobabilistic}: They used known LLM backbones for unified probabilistic time-series forecasting.
\end{itemize} 

To compare with the baselines, we adopt MAE, RMSE, and MAPE prediction error~\cite{li2024flashstsimpleuniversalprompttuning}.

\subsection{Proof of Equation \ref{kl_formula}}\label{kl_proof}

Let $f_d({\bf x}\mid \boldsymbol{\mu},\boldsymbol{\Sigma},\nu)$ be a multivariate $t$, of dimension $d$, with location vector $\boldsymbol{\mu}$, scale matrix $\boldsymbol{\Sigma}$ and $\nu$ degrees of freedom. The aim is to define an objective prior for the parameter $\nu$. For simplicity in the notation, we will write $f_{d,\nu}=f_d({\bf x}\mid \boldsymbol{\mu},\boldsymbol{\Sigma},\nu)$, for $\nu=1,\ldots,\nu_{\max-1}$, and $f_{d,\nu_{\max}}=N_d({\bf x}\mid \boldsymbol{\mu},\boldsymbol{\Sigma})$, with
$$N_d({\bf x}\mid \boldsymbol{\mu},\boldsymbol{\Sigma}) = \frac{1}{\sqrt{(2\pi)^d|\boldsymbol{\Sigma}|}}\exp\left\{-\frac{1}{2}({\bf x}-\boldsymbol{\mu})^\top\boldsymbol{\Sigma}^{-1}({\bf x}-\boldsymbol{\mu})\right\},$$
where in this case $\boldsymbol{\mu}$ is the vector of means and $\boldsymbol{\Sigma}$ is the covariance matrix.
The prior for $\nu$ here discussed depends on the Kullback--Leibler divergence between two multivariate densities. In particular, for $\nu=1,\ldots,\nu_{\max-1}$, the prior is based on the Kullback--Leibler divergence between two multivariate $t$ densities, which differ only in the number of degrees of freedom.
The divergence between two $d$-variate $t$ densities, $f_{d,\nu}$ and $f_{d,\nu^{\prime}}$, is given by
\begin{eqnarray}\label{DKLt}
&&D_{KL}(f_{d}(\cdot\mid \boldsymbol{\mu},\boldsymbol{\Sigma},\nu)\mid\mid f_{d,}(\cdot\mid \boldsymbol{\mu},\boldsymbol{\Sigma},\nu^{\prime})) \\
&=& D_{KL}(f_{d}(\cdot\mid {\bf 0},{\bf I},\nu)\mid\mid f_{d}(\cdot\mid  {\bf 0},{\bf I},\nu^{\prime})) \nonumber\\
&=&\int_{{\mathbb R}^n} f_{d}({\bf x}\mid {\bf 0},{\bf I},\nu) \log \dfrac{f_{d}({\bf x}\mid {\bf 0},{\bf I},\nu)}{f_{d}({\bf x}\mid {\bf 0},{\bf I},\nu^{\prime})} \,d{\bf x}\nonumber\\
&=& \int_{{\mathbb R}^n} K(d,\nu)\left(1+\dfrac{{\bf x}^{\top}{\bf x}}{\nu}\right)^{-\frac{\nu+d}{2}} \log \dfrac{K(d,\nu)\left(1+\dfrac{{\bf x}^{\top}{\bf x}}{\nu}\right)^{-\frac{\nu+d}{2}}}{K(d,\nu^{\prime})\left(1+\dfrac{{\bf x}^{\top}{\bf x}}{\nu^{\prime}}\right)^{-\frac{\nu^{\prime}+d}{2}}} \,d{\bf x}\nonumber\\
&=& \log \dfrac{K(d,\nu)}{K(d,\nu^{\prime})} -\dfrac{\nu+d}{2}{\mathbb E}_{d,\nu}\left[ \log\left(1+\dfrac{{\bf x}^{\top}{\bf x}}{\nu}\right)\right] +\\
&&\dfrac{\nu^{\prime}+d}{2}{\mathbb E}_{d,\nu}\left[ \log\left(1+\dfrac{{\bf x}^{\top}{\bf x}}{\nu^{\prime}}\right)\right],
\end{eqnarray}
where
$$K(d,\nu)=\dfrac{\Gamma\left(\dfrac{\nu+d}{2}\right)}{\Gamma\left(\dfrac{\nu}{2}\right)\sqrt{(\pi\nu)^d}},$$
and ${\mathbb E}_{d,\nu}$ represents the expected value with respect to $f_{d}(\cdot \mid {\bf 0},{\bf I},\nu)$. More specifically,
\begin{eqnarray*}
{\mathbb E}_{d,\nu}\left[ \log\left(1+\dfrac{{\bf x}^{\top}{\bf x}}{\nu}\right)\right]  =  \Psi\left(\dfrac{\nu+d}{2}\right) - \Psi\left(\dfrac{\nu}{2}\right),
\end{eqnarray*}
where $\Psi$ is the digamma function. For the second expectation in \eqref{DKLt}, we use Lemma 2 in \cite{10.1006/jmva.1999.1824} to obtain the following expression after a change of variables in terms of spherical coordinates
% \begin{eqnarray*}
% \small{
% {\mathbb E}_{d,\nu}\left[ \log\left(1+\dfrac{{\bf x}^{\top}{\bf x}}{\nu^{\prime}}\right)\right] = K(d,\nu) \dfrac{\pi^{\frac{d}{2}}}{\Gamma\left(\frac{d}{2}\right)}  \int_0^{\infty} \left(1+\dfrac{t}{\nu}\right)^{-\frac{\nu+d}{2}} t^{\frac{d}{2}-1}\log\left(1+\dfrac{t}{\nu^{\prime}}\right) dt,}
% \end{eqnarray*}
% \begin{eqnarray*}
% \small{
% {\mathbb E}_{d,\nu}\left[\log\left(1+\frac{{\bf x}^{\top}{\bf x}}{\nu^{\prime}}\right)\right] &=& 
% K(d,\nu) \frac{\pi^{\frac{d}{2}}}{\Gamma\left(\frac{d}{2}\right)} \times \\
% &&\int_0^{\infty} \left(1+\frac{t}{\nu}\right)^{-\frac{\nu+d}{2}} 
% t^{\frac{d}{2}-1}\log\left(1+\frac{t}{\nu^{\prime}}\right) dt
% }
% \end{eqnarray*}
\begin{eqnarray*}
\mathbb{E}_{d,\nu}\left[\log\left(1+\frac{\mathbf{x}^{\top}\mathbf{x}}{\nu'}\right)\right] &=& 
K(d,\nu) \frac{\pi^{\frac{d}{2}}}{\Gamma\left(\frac{d}{2}\right)} \times \\
&&\int_0^{\infty} \left(1+\frac{t}{\nu}\right)^{-\frac{\nu+d}{2}} 
t^{\frac{d}{2}-1}\log\left(1+\frac{t}{\nu'}\right) dt
\end{eqnarray*}
which only requires one-dimensional numerical integration, regardless of the dimension $d$.

\begin{figure*}[!t]
\centering
\includegraphics[width=0.8\textwidth, keepaspectratio]{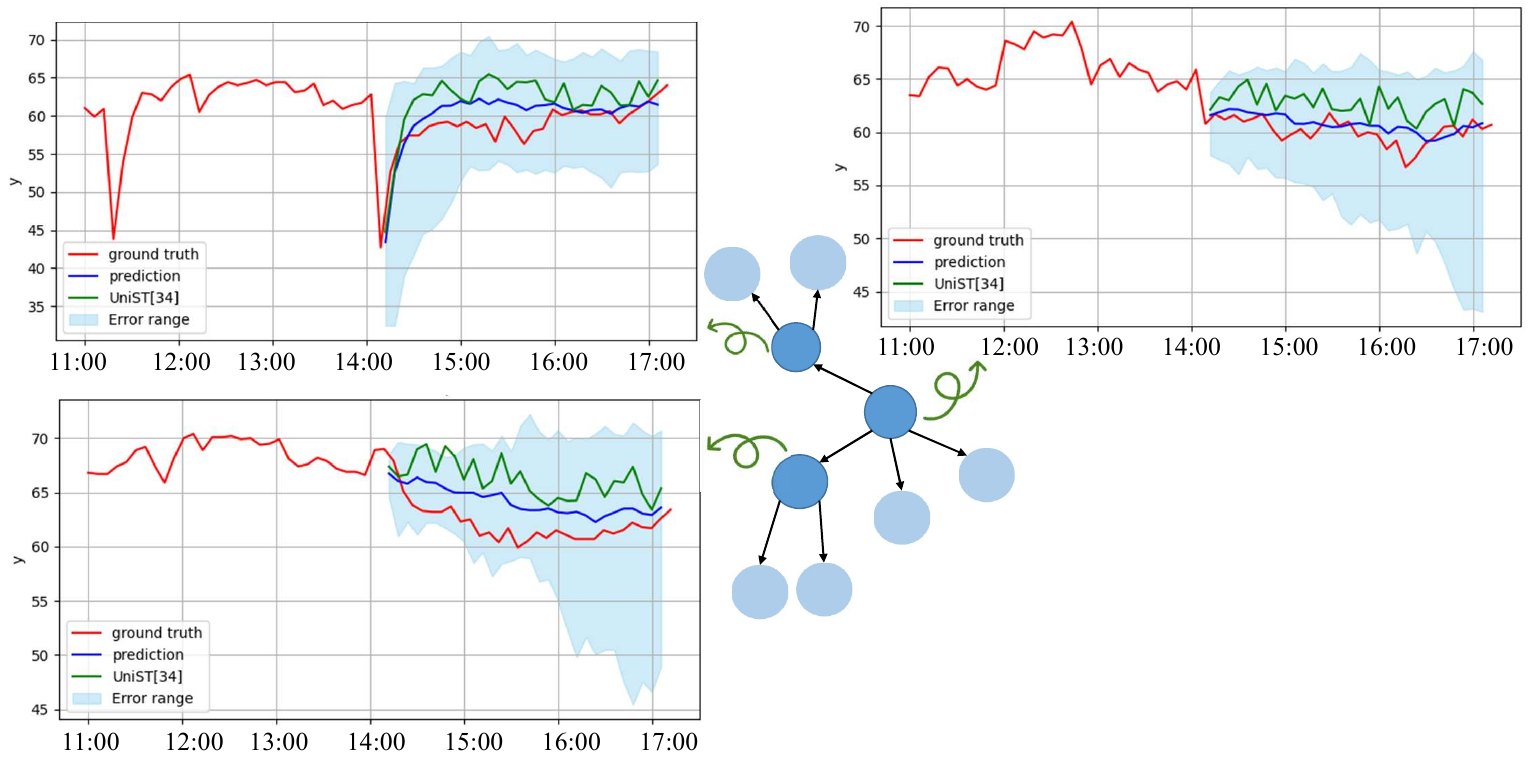}
% \vspace{-3cm}
\caption{The visualization of $k$-hop multivariate zero-shot prediction for Pems07(M) dataset for long-term horizon
}
\label{fig_vis_graph}
\end{figure*}

\begin{figure*}[h]
    \centering
    \begin{minipage}[b]{1.0\textwidth}
        \centering
        \subfigure[Node 6]{
            \includegraphics[width=0.40\textwidth,keepaspectratio]{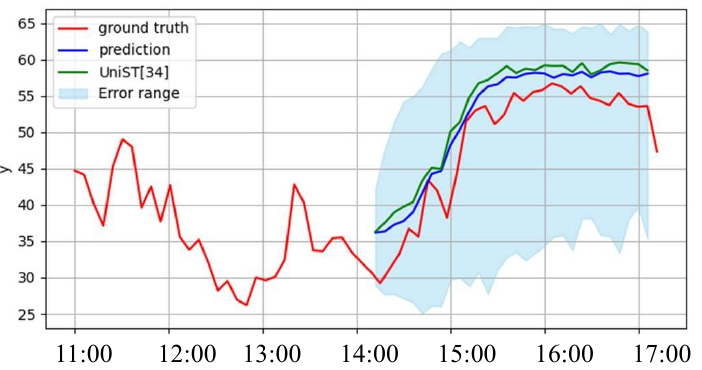}
            \label{fig-2-aa}
        }
        \hfill
        \subfigure[Node 13]{
            \includegraphics[width=0.40\textwidth,keepaspectratio]{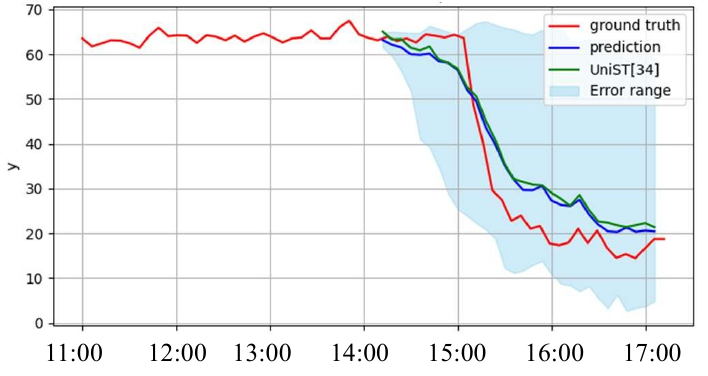}
            \label{fig-2-bb}
        }
    \end{minipage}
    
    \vspace{1em}
    
    \begin{minipage}[b]{1.0\textwidth}
        \centering
        \subfigure[Node 48]{
            \includegraphics[width=0.40\textwidth,keepaspectratio]{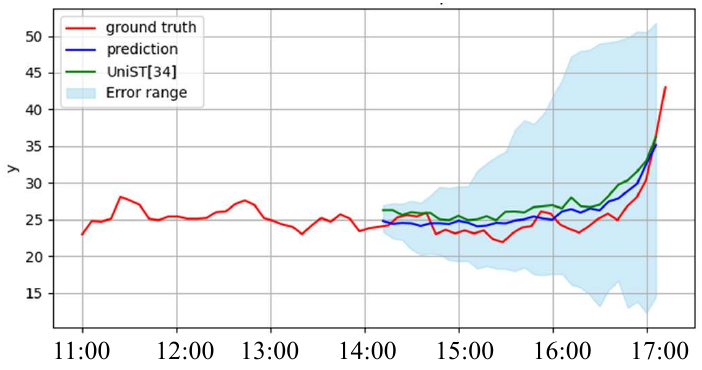}
            \label{fig-2-bc}
        }
    \end{minipage}
    \caption{Strada-LLM traffic prediction in the capturing volatility scenarios for three (not-connected) different nodes exist in Pems07(M) dataset.}
    \label{fig:fig_vol}
\end{figure*}

\subsection{Visualizations}~\label{vis}

To better understand and assess our model visually, we present additional illustrations in this section. The visualization of long-term forecasting in zero-shot setting for Pems07(M) dataset is shown in Figure~\ref{fig_vis_graph}. Figure~\ref{fig_vis_graph} is an instance of multivariate probabilistic forecasting in which the dark blue nodes are part of the k-hop extracted subgraph. 
A critical aspect of time-series prediction algorithms is their ability to capture volatility importance - the varying degrees of uncertainty and rapid changes in the data over time. By incorporating volatility measures, these algorithms can better understand periods of high traffic turbulence versus relative stability, leading to more nuanced and accurate predictions. Strada-LLM is also able to capture volatility existing in Pems07(M) dataset, which is shown in Figure ~\ref{fig:fig_vol}.

\subsection{Gradient Analysis}~\label{grad_vis}
\begin{figure*}[h]
    \centering
    \subfigure[Comparison of $l_2$ gradient norms between the first layer (left) and the second layer (right) across training iterations.]{
        \centering
        \includegraphics[width=0.6\linewidth,keepaspectratio]{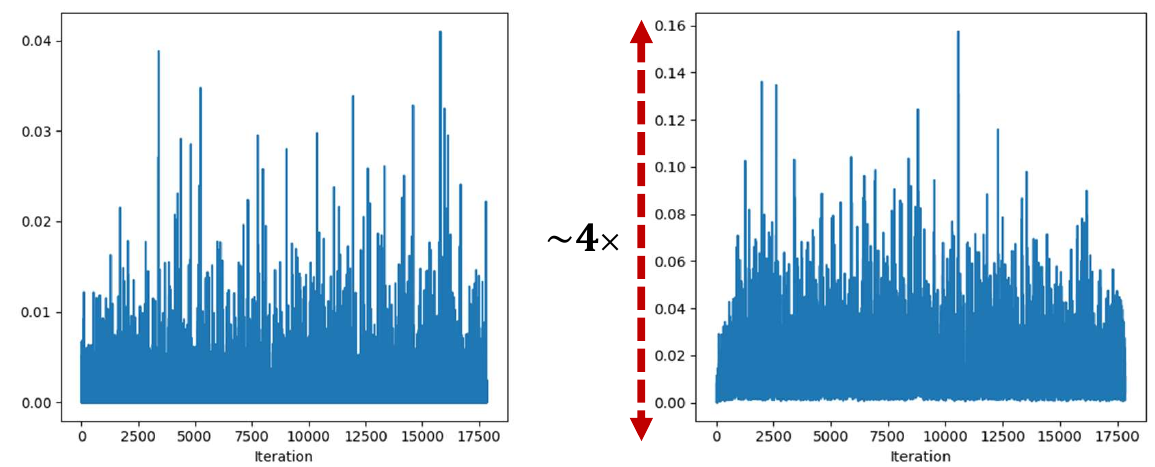}
        % \caption{Strada-LLM}
        \label{fig:fig-grad-norm}
    }
    \hfill
    \subfigure[Gradient $\sigma^2$ variance comparison between LaplacianPE1 (left) and LaplacianPE2 (right) layers during training.]{
        \centering
        \includegraphics[width=0.6\linewidth,keepaspectratio]{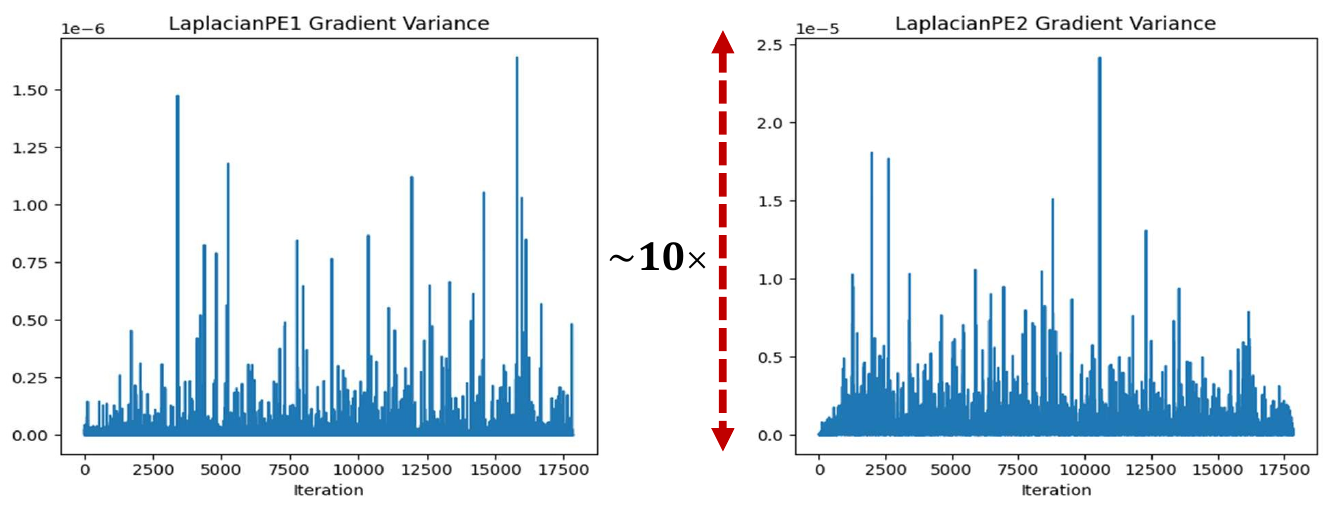}
        % \caption{UniST\cite{unist}}
        \label{fig-2-b-var}
    }
    \caption{Training dynamics comparison between two graph Laplacian positional encoding layers: gradient norms and variances over iterations.}
    \label{fig:fig-grad-var}
\end{figure*}

As mentioned in section ~\ref{domain}, the low-rank adaptation has not been utilised in the Graph Embedding module,  which causes model performance degradation. To explain the reason, the norm and variance magnitude of the gradients of two MLP layers inside the module are plotted in Figure~\ref {fig:fig-grad-var}. As shown in Figure~\ref{fig-2-b-var}, according to the heterogeneous gradient backpropagation of these layers, e.g., one MLP's variance is 10 times the other MLP layer, and the adoption of the same low-rank adaptation causes performance degradation. Proposing an efficient low-rank adaptation in this use case will remain an ongoing research direction.  

\subsection{{Training Strategy}}\label{strategy}

\subsubsection{Augmentation Techniques}
We adopted the time series augmentation techniques $Freq\-Mix$\cite{chen2023fraug} and
$Freq\-Mask$ \cite{chen2023fraug} to prevent over-fitting. Our training framework comprises two distinct stages: pre-training and fine-tuning. During pre-training, we leverage a diverse combination of datasets: $D_{base}$ for general language understanding, $D_{domain}$ for domain-specific knowledge, and $D_{task}$ for task-oriented capabilities. These datasets are processed in a curriculum learning manner, starting with $D_{base}$ to establish foundational knowledge, followed by $D_{domain}$ to inject domain expertise, and finally $D_{task}$ to enhance task-specific performance. To maintain balanced exposure across datasets while optimizing computational efficiency, we employ a dynamic batch construction strategy where each training batch $B_t$ consists of samples drawn from multiple datasets with proportional representation: $\alpha_1$\% from $D_{base}$, $\alpha_2$\% from $D_{domain}$, and $\alpha_3$\% from $D_{task}$, where $\sum_i \alpha_i = 1$. This balanced approach ensures the model develops robust general capabilities while maintaining strong performance on domain-specific tasks. In the fine-tuning stage, we exclusively focus on $D_{task}$ with smaller learning rates and specialized optimization techniques to prevent catastrophic forgetting of pre-trained knowledge while adapting to target objectives.

\subsubsection{Experimental Setup}\label{experimental_setup}
The hyperparameters of the Strada-LLM model mainly include:
learning rate, batch size, training epochs, the number of layers, context length, embedding dimension per head, number of heads, rope scaling, and number of parallel samples for greedy decoding. We adopt Adam as the optimizer~\cite{kingma2014adam}. In the experiment, we follow the weight decay $1e-8$ and set the learning rate to 0.001~\cite{kingma2014adam}, the batch size to 32, and the training
epochs to 150 for each dataset. We set $k=3$ in the sub-graph extractor. Furthermore, based on the fact that Strada-LLM is a probabilistic model, we take samples and compute medians to compare with point-wise models. The number of samples is set to $100$. In terms of complexity, our model contains 16 million parameters. The experiments are conducted on 2 Nvidia RTX 4090 GPUs. 

\begin{table}[htbp]
\caption{LoRA rank benchmark results on PeMS04 dataset}
\begin{center}
\begin{tabular}{lccc}
\toprule
$R$ & MAE$\downarrow$ & RMSE$\downarrow$ & MAPE(\%)$\downarrow$ \\
\midrule
$2$ & 24.76 & 29.37  & 12.80 \\
$4$ & \underline{22.12} & \underline{27.58}  & \underline{12.09} \\
$8$ & \textbf{21.87} & \textbf{26.62}  & \textbf{11.8} \\
\bottomrule
\end{tabular}
\label{tab_rank}
\end{center}
\end{table}

\subsection{Rank Analysis}\label{rank_analyse}
% \vspace{-0.3cm}
We evaluate the performance of our LoRA adoption with different widths. To do so, we try various values for $r \in (2, 4, 8)$ and the results are demonstrated in Table~\ref{tab_rank}. As can be seen, the lower error comes with the cost of involving more parameters. For real-time applications, the lower bandwidths are more applicable.

\subsection{Perturbation Analysis and Robustness}\label{robust}

\begin{figure}[!htb]
\includegraphics[scale=0.3]{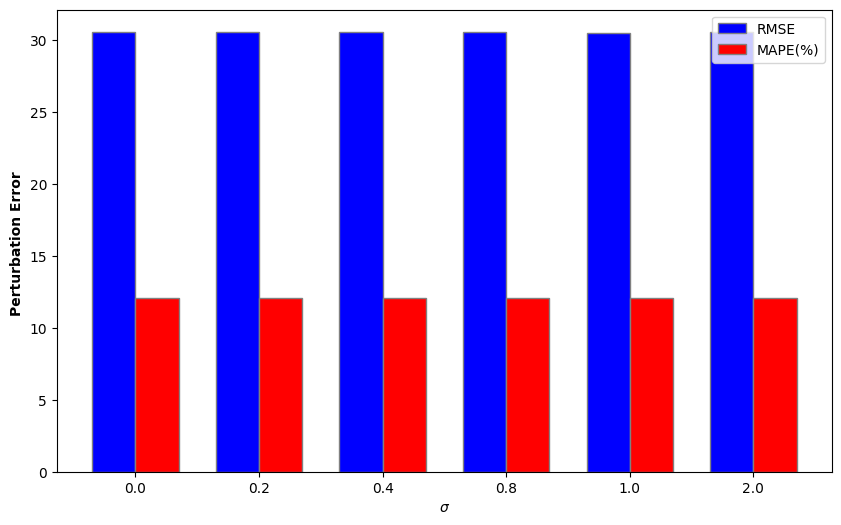}
\centering\caption{Evaluation metrics for the perturbation analysis. Different $\sigma$ values were examined to test the model's robustness.}
% \centering\caption{proposed LoRA method for transfer learning}
\label{fig4p}
\end{figure}

To evaluate the robustness of the Strada-LLM model against noise, we conduct perturbation analysis experiments. We added two types of common random noise to {the validation data}
during the experiment. Random noise obeys the Gaussian distribution $ \mathcal{N}(0, \sigma^2)$ where $\sigma \in (0.2, 0.4, 0.8, 1, 2)$. Then, we normalize the values of the noise matrices to be between 0 and 1. The results are shown in Figure~\ref{fig4p}, where the horizontal axis represents $\sigma$, the vertical axis represents the error, and different colors indicate different evaluation metrics. This benchmark is evaluated for the dataset PeMS04 but other datasets follow the same pattern. According to this experiment, our model is robust to Gaussian noise.

\begin{figure}[!htb]
\includegraphics[scale=0.3]{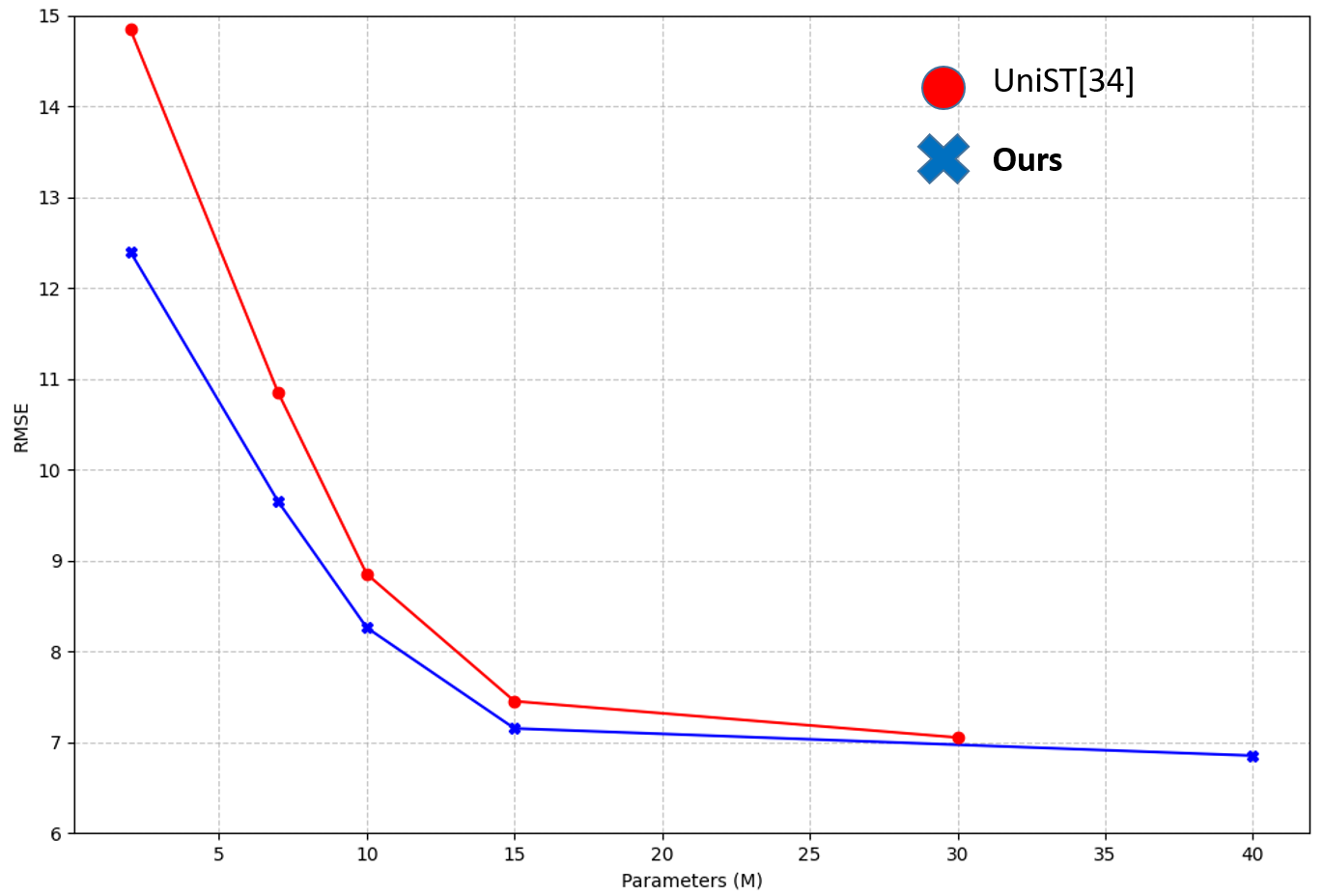}
\centering\caption{Strada-LLM scalability analysis compared to existing LLMs.}
% \centering\caption{proposed LoRA method for transfer learning}
\label{fig10p}
\end{figure}

\subsection{Scalability Comparison}\label{scalability}
Our empirical analysis reveals a nuanced relationship between model size and performance, characterized by diminishing returns beyond certain parameter thresholds. As illustrated in Figure~\ref{fig10p}, our model performs superior on the similar parameter capacity on the Crowd dataset.

\end{document}